\documentclass[lettersize,journal]{IEEEtran}
\ifCLASSOPTIONcompsoc
  \usepackage[nocompress]{cite}
\else
  \usepackage{cite}
\fi
\ifCLASSINFOpdf
   \usepackage[pdftex]{graphicx}
   \graphicspath{{../pdf/}{../jpeg/}}
\else
   \usepackage[dvips]{graphicx}
   \graphicspath{{../eps/}}
\fi
\usepackage{booktabs}
\usepackage{mathrsfs}
\usepackage{amsmath}
\usepackage{amssymb}
\usepackage{amsfonts}
\usepackage{fdsymbol}
\usepackage{diagbox}
\usepackage{soul, color, xcolor}

\usepackage{arydshln}

\usepackage{tabu}
\usepackage{algorithmic}
\usepackage{algorithm}


\usepackage{array}

\ifCLASSOPTIONcompsoc
  \usepackage[caption=false,font=footnotesize,labelfont=sf,textfont=sf]{subfig}
\else
  \usepackage[caption=false,font=footnotesize]{subfig}
\fi
\usepackage{multirow}
\usepackage{caption}

\usepackage{stfloats}

\usepackage{color}

\usepackage{url}
\usepackage{ulem}

\newtheorem{definition}{Definition}[section]

\begin{document}
%
\title{GLOW: Graph-Language Co-Encoding for Agentic Workflow Performance Prediction}
%
%
%
%

\author{
Wei Guan,
Jian Cao,~\IEEEmembership{Senior Member,~IEEE},
Jinyu Cai,
Qiqi Cai,
Jianqi Gao,
See-Kiong Ng,~\IEEEmembership{Member,~IEEE}
\IEEEcompsocitemizethanks{
\IEEEcompsocthanksitem This work was supported by the China National Science Foundation (Grant No. 62072301) and the Smart Medical Special Research Fund of the Shanghai Municipal Health Commission (Grant No. 2025ZHYL003).
 (Corresponding author: Jian Cao.)
\IEEEcompsocthanksitem 
W. Guan is with the School of Computer Software, Tianjin University, Tianjin, China. E-mail: guan\_wei@tju.edu.cn.
\IEEEcompsocthanksitem 
J. Cao, and Q. Cai are with the School of Computer Science, Shanghai Jiao Tong University, China. 
E-mail: \{cao-jian, cai\_qiqi\}@sjtu.edu.cn.
\IEEEcompsocthanksitem 
J. Cai and S.-K. Ng are with the Institute of Data Science, National University of Singapore, Singapore. 
E-mail: \{jinyucai, seekiong\}@nus.edu.sg.
\IEEEcompsocthanksitem 
J. Gao is with the School of Computer Engineering and Science, Shanghai University, China. 
E-mail: jianqi\_gao@shu.edu.cn.
}
}

\markboth{ }%
{Shell \MakeLowercase{\textit{et al.}}: Bare Demo of IEEEtran.cls for Computer Society Journals}
%



\IEEEtitleabstractindextext{%
\begin{abstract}
Agentic Workflows (AWs) have emerged as a promising paradigm for solving complex tasks.
However, automatically generating high-quality AWs remains expensive because AW optimization requires evaluating a large number of candidate AWs via execution, resulting in high computational cost and latency.
Recently, AW performance prediction has become a hot research topic to avoid costly execution-based evaluation, but existing methods primarily use Graph Neural Networks (GNNs) to model workflow structures and insufficiently capture the semantic relationships among agents.
To address this limitation, we propose GLOW, a unified framework for AW performance prediction that combines the graph-structure modeling ability of GNNs with the topology-aware semantic encoding capability of LLMs. 
Specifically, a graph-oriented LLM is first built through instruction-tuning on graph understanding tasks to extract topology-aware semantic representations from descriptive text of AWs.
Meanwhile, a GNN explicitly models the structural information of AWs and produces corresponding structural representations.
The semantic and structural representations are then fused in a shared latent space using a Transformer-based fusion module. 
A contrastive learning strategy is further introduced to learn more discriminative representations for AWs.
Experiments on the FLORA-Bench benchmark demonstrate that GLOW consistently outperforms state-of-the-art baselines in both prediction accuracy and ranking utility. 
Moreover, when integrated into the AFLOW, an automatic AW generation framework, GLOW reduces optimization time by 98.7\% with only a 0.031 average score decrease across three datasets, showing its effectiveness as an efficient surrogate evaluator for AW optimization.
\end{abstract} 

}

\maketitle

\IEEEdisplaynontitleabstractindextext

%
\IEEEpeerreviewmaketitle

\section{Introduction}\label{sec:introduction}
\IEEEPARstart{L}{arge} Language Models (LLMs) have demonstrated remarkable capabilities across a wide range of tasks, evolving from passive text generators to active agents capable of planning, reasoning, and tool use \cite{xi2025rise}.
Recently, there has been growing interest in building Agentic Workflows (AWs), where LLM-based agents are orchestrated through structured interaction topologies to collaboratively solve complex tasks \cite{dawid2025agentic}. In this paradigm, agents are organized into workflow structures that define roles (e.g., planner, executor, verifier, or critic) and govern multi-step reasoning, tool use, and information exchange. 
Compared to conventional single-step or unstructured prompting strategies, AWs provide a more modular, controllable, and scalable framework by decomposing complex problems into sub-tasks executed within structured processes, enabling intermediate reasoning, self-reflection, and cross-step verification to improve reliability and performance \cite{moncada2025agentic}. 
As a result, AWs have achieved promising performance in a variety of challenging domains, including code generation \cite{he2025llm,hu2025self,liu2025sew,watanabe2025use}, mathematical reasoning \cite{zhong2026achieving,zhang2025debate4math,gou2024tora}, and general problem solving \cite{pezeshkpour2024reasoning,chen2025magicore,wang2026dyflow}.

However, designing effective AWs manually is labor-intensive and requires expert knowledge, which has motivated the development of automatic AW generation methods \cite{zhuge2024gptswarm,zhang2024g,zhang2024aflow,tong2026wirelessagent,li2024autoflow,hu2025self,hu2024automated,xu2025robustflow,wang2026dyflow,liu2025sew,niu2025flow,zhao2026a2flow,zhou2025agentic,yang2025autoflowgen,tan2025meta}.  
These methods typically formulate automatic AW generation as an iterative optimization problem, where candidate AWs are explored via search or policy optimization and refined using performance feedback from task-specific evaluations.
Nevertheless, a critical bottleneck that impedes their scalability lies in the evaluation of candidate AWs. In  existing approaches, the performance of an AW is obtained through full end-to-end execution, where each agent invokes an LLM to complete its assigned subtask and passes intermediate outputs to downstream agents according to the workflow topology.
Given the substantial computational overhead of repeated LLM calls and the inherent complexity of multi-turn interactions, the evaluation process becomes both time-consuming and expensive, thereby hindering large-scale exploration of candidate AWs.

To address this issue, recent works  \cite{zhang2025gnns,trirat2025agentic}  have explored performance predictors as surrogates for execution-based evaluation. 
These methods model AWs as Directed Acyclic Graphs (DAGs) and utilize Graph Neural Networks (GNNs) to predict performance based on structural features. While effective at capturing topological patterns, standard GNNs treat agent prompts as non-contextualized text embeddings, often failing to comprehend the deep semantic logic and role definitions critical to AW success.
On the other hand, although LLMs, such as Qwen3 \cite{yang2025qwen3}, LLaMA 4 \cite{adcock2026llama}, and DeepSeek-R1 \cite{guo2025deepseek}, are powerful at semantic understanding, directly using them to predict the performance of AWs is  suboptimal.
They tend to treat AWs as linear text and thus capture graph structure, such as dependencies and branching, only to a limited extent. More importantly, they struggle to model error propagation across agents, where mistakes made by upstream agents may cascade through the workflow and ultimately affect the final outcome. As a result, their predictions often rely more on superficial semantics than execution behavior, leading to unreliable performance estimates.

Considering that both structural information (i.e., interaction topology over workflow nodes) and semantic information (i.e., node-level prompts and their dependency-aware execution behaviors) are essential for AW performance prediction, we propose \textbf{GLOW}, a unified framework that leverages the structural modeling capabilities of \textbf{G}NNs and the semantic encoding power of \textbf{L}LMs for agentic workfl\textbf{OW} performance prediction. 
A graph-oriented LLM is first constructed through instruction-tuning on a suite of graph understanding tasks (e.g., reachability, topological sorting), enabling it to extract topology-aware semantic representations from descriptive text of AWs.
On this basis, a dual-branch architecture is designed for AW representation encoding.
Specifically, a GNN branch explicitly performs message passing to learn structural representations that capture dependencies and multi-hop information propagation among nodes.
In parallel, the graph-oriented LLM branch provides complementary semantic representations of AWs. 
Importantly, the graph-oriented LLM takes a descriptive text of AW (graph) as input, including both node prompts and edge relations. Instead of performing explicit message passing over the graph, it encodes topology-aware semantic representations holistically. 
The two branches produce complementary structural and semantic representations of the AW, which, together with the task instruction representation, are projected into a unified latent space and fused via a transformer-based representation fusion module for performance prediction.
During training, in addition to the supervision signal for performance prediction, a contrastive learning objective is introduced to further improve discriminability, which encourages successful AWs for a given task to cluster together while separating them from unsuccessful ones in the latent space.
Extensive experiments on the FLORA-Bench benchmark demonstrate that GLOW consistently outperforms existing state-of-the-art baselines in both prediction accuracy and ranking utility. 
Furthermore, when used as a surrogate evaluator to replace execution-based AW evaluation in the automatic AW generation framework AFLOW \cite{zhang2024aflow}, GLOW reduces computation time by 98.7\% while incurring only a 0.031 decrease in score on average across three datasets, outperforming other performance prediction methods with significantly smaller degradation, demonstrating its effectiveness as an efficient surrogate evaluator for AW optimization.
The main contributions of this work are summarized as follows:
\begin{itemize}
    \item We propose GLOW, a unified framework for AW performance prediction, which adopts a dual-branch AW representation encoding architecture consisting of a GNN and a graph-oriented LLM to jointly model AW structure and semantic logic.
    \item We introduce a contrastive learning strategy that improves AW representation quality by pulling representations of successful AWs closer while pushing them away from those of unsuccessful ones in the latent space.
    \item Extensive experiments on FLORA-Bench demonstrate that GLOW achieves state-of-the-art performance. Ablation studies further confirm the effectiveness of the dual-branch AW representation encoding architecture.
\end{itemize}

The remainder of this paper is organized as follows: Section \ref{sec:Relatedworks} reviews related work. Section \ref{sec:Preliminaries} defines the problem formulation. Section \ref{sec:method} introduces the proposed GLOW method. Section \ref{sec:exp} presents the experimental results. Finally, Section \ref{sec:conclusions} concludes the paper and discusses future work.

\begin{table*}[t]
\renewcommand{\arraystretch}{1}
\setlength{\tabcolsep}{2mm}
\caption{A brief summary of related works.}
\label{tab:relatedWork}
\vskip -0.05in
\begin{tabular}{cccc}
\toprule
\multicolumn{2}{c}{Categories} & Methods & Comments/Limitations \\ 
\midrule

\multirow{3}{*}{\begin{tabular}[c]{@{}c@{}}
Automatic Agentic\\
Workflow Generation
\end{tabular}}
&
\begin{tabular}[c]{@{}c@{}}
Probability-Based
\end{tabular}
&
\cite{zhuge2024gptswarm,zhang2024g}
&
\begin{tabular}[c]{@{}c@{}}
Generate AWs through stochastic sampling from a learnable distribution.
\end{tabular}
\\
\cmidrule{2-4}

&
\begin{tabular}[c]{@{}c@{}}
LLM-Guided
\end{tabular}
&
\begin{tabular}[c]{@{}c@{}} 
\cite{hu2025self,liu2025sew,wang2026dyflow}\\
\cite{zhang2024aflow,tong2026wirelessagent,li2024autoflow,hu2024automated,xu2025robustflow,niu2025flow,zhao2026a2flow,zhou2025agentic,yang2025autoflowgen,tan2025meta}
\end{tabular}
&
\begin{tabular}[c]{@{}c@{}}
Leverage reasoning and coding capabilities of LLMs to \\
directly generate and refine AWs based on feedback.
\end{tabular}
\\

\midrule

\multirow{4}{*}{\begin{tabular}[c]{@{}c@{}}
LLMs for\\
Graph-Structured Data
\end{tabular}}
&
\begin{tabular}[c]{@{}c@{}}
Prompt-Based QA
\end{tabular}
&
\cite{ICLR2024_bf72f65f,ye2024language,zhang2024llm4dyg,wang2024instructgraph,zhu2024investigating}
&
\begin{tabular}[c]{@{}c@{}}
Represent graphs as natural language descriptions\\
to enable LLM-based graph understanding and QA.
\end{tabular}
\\
\cmidrule{2-4}

&
\begin{tabular}[c]{@{}c@{}}
GNN-Enhanced QA
\end{tabular}
&
\cite{chai2023graphllm,liu2024can,tang2024graphgpt,wang2024llms}
&
\begin{tabular}[c]{@{}c@{}}
Integrate GNN-derived structure-aware node embeddings with LLMs. 
\end{tabular}
\\

\cmidrule{2-4}
&
\begin{tabular}[c]{@{}c@{}}
Graph Representation Learning
\end{tabular}
&
\cite{xu2025llm}
&
\begin{tabular}[c]{@{}c@{}}
Leverage LLMs as graph encoders to generate graph-level representations.
\end{tabular}
\\

\midrule
\multicolumn{2}{c}{\begin{tabular}[c]{@{}c@{}}
Agentic Workflow\\
Performance Prediction
\end{tabular}}
&
\cite{zhang2025gnns,trirat2025agentic}
&
\begin{tabular}[c]{@{}c@{}}
Lack explicit modeling of topology-aware semantic relationships\\
among agents and adaptive fusion of heterogeneous representations.
\end{tabular}
\\

\bottomrule
\end{tabular}
\vskip -0.1in
\end{table*}

\section{Related Work}
\label{sec:Relatedworks}
In this section, we briefly review prior research on automatic agentic workflow generation, LLMs for graph-structured data, and agentic workflow performance prediction.
Table~\ref{tab:relatedWork} provides an overview of existing studies.

\subsection{Automatic Agentic Workflow Generation}
Current approaches for automated AW generation generally fall into two primary categories.
\textbf{Probability-based} methods \cite{zhuge2024gptswarm,zhang2024g} generate candidate AWs through stochastic sampling from a learnable distribution. 
To facilitate this mathematical optimization, these approaches typically model the AW as a computational graph, where nodes represent LLM-based agents and edges define their interaction topology. 
For example, GPTSwarm \cite{zhuge2024gptswarm} utilizes the REINFORCE algorithm to optimize this graph structure, learning the probability of connections between nodes to maximize the AW performance. 
G-Designer \cite{zhang2024g} employs a variational graph auto-encoder (VGAE) to sample and decode task-adaptive AWs.
\textbf{LLM-guided} methods \cite{zhang2024aflow,tong2026wirelessagent,li2024autoflow,hu2025self,hu2024automated,xu2025robustflow,wang2026dyflow,liu2025sew,niu2025flow,zhao2026a2flow,zhou2025agentic,yang2025autoflowgen,tan2025meta}, conversely, leverage the inherent reasoning and coding capabilities of LLMs to directly generate and refine AWs based on feedback.
For example, AFLOW \cite{zhang2024aflow} utilizes Monte Carlo Tree Search (MCTS) to explore different candidate AWs.
WirelessAgent++ \cite{tong2026wirelessagent} extends this paradigm to the wireless domain, treating AWs as executable programs and employing a domain-adapted MCTS. 
AutoFlow \cite{li2024autoflow} frames AWs as natural language programs, employing reinforcement learning to fine-tune the generator LLM based on AW execution rewards. 
RobustFlow \cite{xu2025robustflow} executes multiple AW candidates for similar user queries, identifies the one that performs the best, and trains the LLM to consistently generate that high-quality AW. 
EvoMAC \cite{hu2025self} mimics neural network training by introducing ``textual backpropagation," where error logs from compilers serve as gradients to update the AWs. 
ADAS \cite{hu2024automated} employs a meta-agent to iteratively program, evaluate, and refine complex AWs, leveraging an evolving archive of discovered designs as stepping stones.

However, these approaches rely heavily on repeated LLM invocations to execute AWs for performance evaluation, resulting in substantial computational, temporal, and financial overhead, which limits their practicality in real-world scenarios. 
GLOW provides an efficient way to predict the performance of generated candidate AWs, thereby reducing the need for costly LLM calls.

\subsection{LLMs for Graph-Structured Data}
A growing body of work has investigated the use of LLMs for graph understanding and Question Answering (QA). Wang et al. \cite{wang2023can} introduce one of the first natural-language graph understanding benchmarks, NLGraph, and demonstrate that LLMs exhibit promising capabilities in understanding and reasoning over graph-structured data. 
Building upon such benchmarks, existing approaches can be broadly categorized into two groups.
\textbf{Prompt-based} methods \cite{ICLR2024_bf72f65f,ye2024language,zhang2024llm4dyg,wang2024instructgraph,zhu2024investigating} represent graphs in natural language formats and leverage prompt engineering or instruction tuning to elicit the capabilities of LLMs on graph-related tasks.
\textbf{GNN-enhanced} methods \cite{chai2023graphllm,liu2024can,tang2024graphgpt,wang2024llms}, in contrast, further enhance this paradigm by integrating GNN-derived structure-aware node embeddings with LLMs. 
Specifically, structure-aware node embeddings are incorporated into prompts to better align graph topology information with LLMs' semantic space, thereby improving their ability to reason over graph-structured data.

In contrast to these approaches, GLOW does not use LLMs for graph-specific question answering. Instead, GLOW leverages LLMs as representation encoders to produce richer semantic representations of AWs, which serve as inputs for downstream performance prediction. 
Along this line, Xu et al. \cite{xu2025llm} propose GALLON, which leverages off-the-shelf LLMs to produce graph-level representations for molecular graphs, demonstrating that LLMs can serve as powerful encoders for capturing the semantics of graph-structured data.
Notably, our work goes beyond the use of off-the-shelf LLMs. 
A graph-oriented LLM is introduced via instruction-tuning on graph understanding tasks, enabling better encoding of topology-aware semantic and functional information in AWs.

\subsection{Agentic Workflow Performance Prediction}
To mitigate the prohibitive cost of evaluating AWs via direct execution, recent research has shifted towards developing lightweight performance predictors. 
Zhang et al. \cite{zhang2025gnns} pioneered this direction by formulating AWs as DAGs and leveraging GNNs to capture AW topology and derive graph-level representations. The performance is then predicted using a Multi-Layer Perceptron (MLP) based on the concatenation of the AW representation and the task representation.
Subsequently, Trirat et al. \cite{trirat2025agentic} introduced Agentic Predictor, which extends the graph-based AW encoding paradigm by incorporating multiple workflow views, including graph structures, code implementations, and prompts.
Specifically, it employs GNNs to encode graph structures, pretrained embedding models to encode code implementations and prompts, and concatenates the extracted representations for downstream performance prediction.

However, existing AW performance predictors mainly rely on conventional encoders and concatenation-based fusion strategies, which may limit their ability to explicitly model topology-aware semantic relationships among agents and adaptively fuse  heterogeneous representations.
In contrast, GLOW introduces a graph-oriented LLM and a GNN to jointly capture the semantic and structural characteristics of AWs. 
The graph-oriented LLM is instruction-tuned on graph understanding tasks to learn topology-aware semantic representations, while the GNN encodes AW structural information. 
Moreover, instead of relying on concatenation-based fusion, GLOW employs a Transformer-based fusion module, where the attention mechanism adaptively fuses heterogeneous representations by learning the interactions and relative importance among different representations.
Furthermore, GLOW incorporates a contrastive learning strategy to refine the learned representations. Specifically, for a given task, successful and unsuccessful AWs are pushed apart in the latent space, resulting in more discriminative representations for AW performance prediction.

\begin{figure}[tb]
	\centering
    \includegraphics[scale=1]{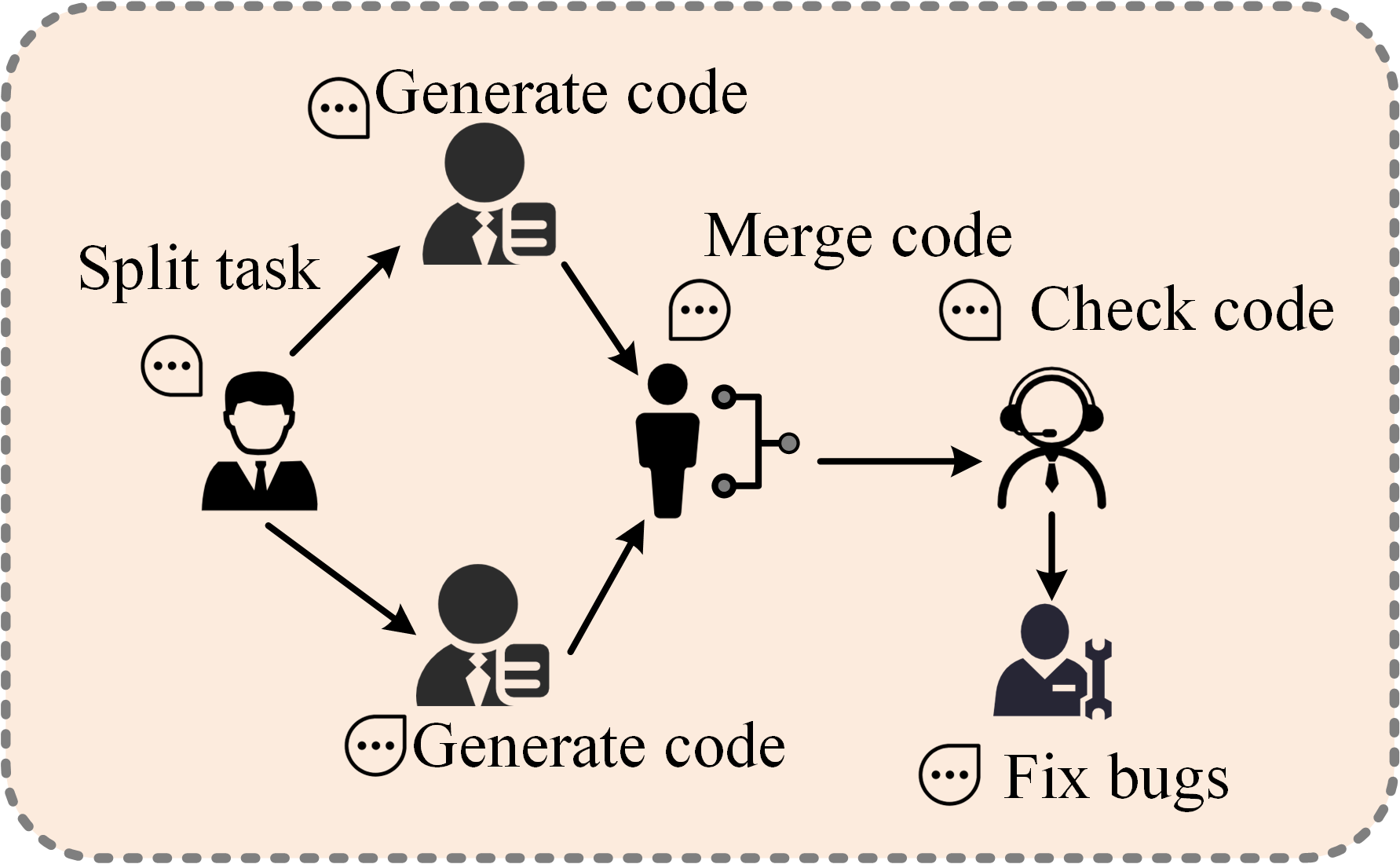}
    	\vskip -0.05in
 	\caption{An illustrative example of an AW  for code generation.}
	\label{fig:af_example}
    \vskip -0.1in
\end{figure}

\begin{figure*}[tb]
	\centering
    \includegraphics[width=\textwidth]{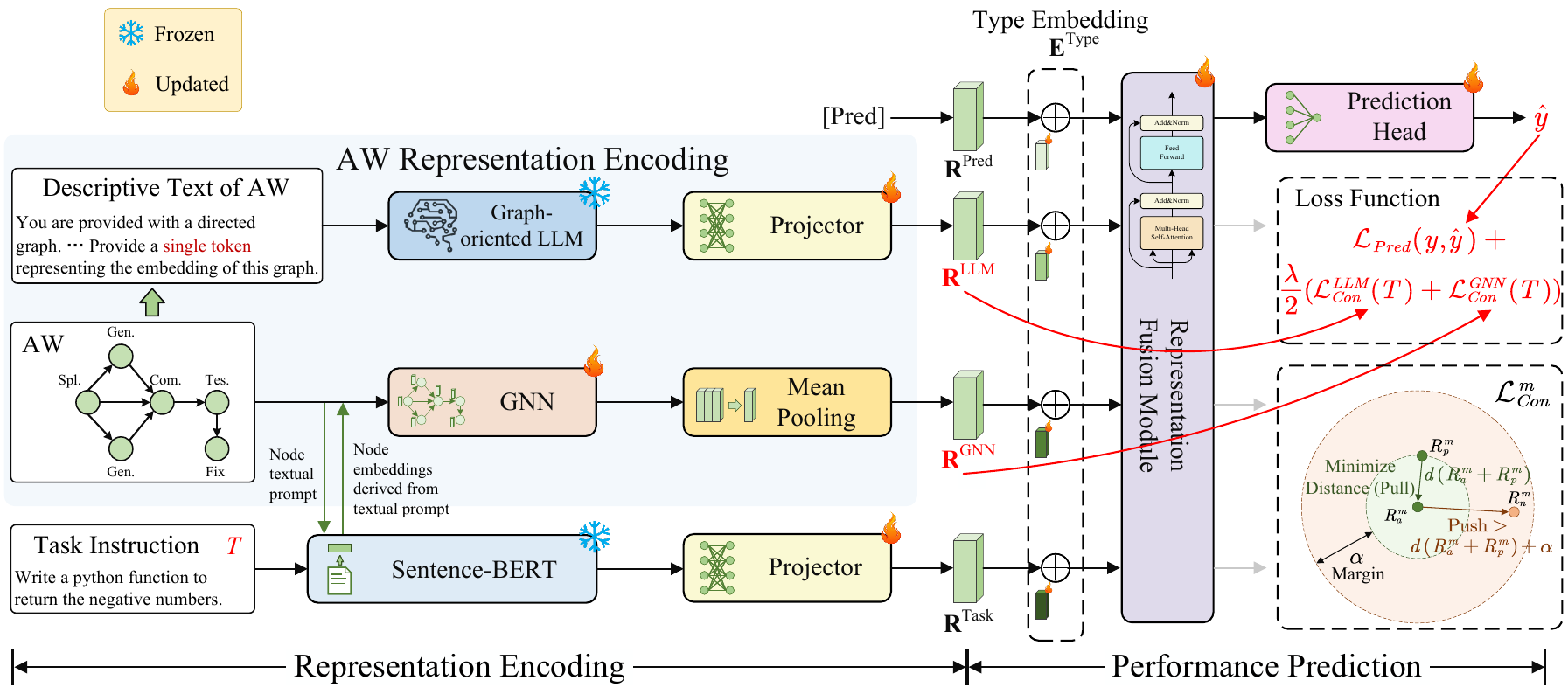}
 	\caption{The architecture of the proposed GLOW and the parameter optimization status during end-to-end model training. `Frozen' denotes modules whose parameters are fixed, while `Updated' denotes modules optimized through backpropagation during this stage.
    }
	\label{fig:architecture}
    \vskip -0.1in
\end{figure*}

\section{Preliminaries}
\label{sec:Preliminaries}
An \textit{Agentic Workflow} (AW) consists of multiple collaborating LLM-based agents, each associated with a prompt that defines its behavior, that collectively execute a task $T$ by passing information, triggering actions, and maintaining interdependent states.
As illustrated in Fig. \ref{fig:af_example}, such AWs typically exhibit structured control flow and explicit dependency relationships among agents.
To formally characterize these interaction patterns, an AW is abstracted as a Directed Acyclic Graph (DAG), following previous studies \cite{zhang2025gnns,trirat2025agentic}. Specifically, an AW with $N$ agents is represented as $\mathcal{G} = \{\mathcal{V}, \mathcal{E}, \mathcal{P}\}$, where $\mathcal{V} = \{v_1, v_2, \dots, v_N\}$ denotes the set of agents. The edge set $\mathcal{E}$ captures the directional dependency relationships between agents, representing information and execution flow.
The prompt set $\mathcal{P} = \{p_1, p_2, \dots, p_N\}$ contains the textual prompts associated with each agent, where $p_i$ defines the behavior of agent $v_i$.
During the AW \textit{execution} phase, each agent $v_i$ aggregates information from two sources: the initial global task instruction $T$ and the intermediate outputs generated by its upstream neighbors. The input context $X_i$ for agent $v_i$ can be expressed as:
\begin{equation}
    X_i = \{T\} \cup \{y_j \mid v_j \in \mathcal{N}^{(in)}_i\}
    \label{eq:input_formulation}
\end{equation}
where $\mathcal{N}^{(in)}_i$ signifies the set of predecessor agents (nodes) directly connected to $v_i$, and $y_j$ represents the output produced by agent $v_j$. 
Based on this input context, the output $y_i$  for agent $v_i$  is generated by invoking LLMs, denoted as $\mathcal{M}$. The generation process is defined by:
\begin{equation}
    y_i = \mathcal{M}(X_i, p_i)
    \label{eq:output_generation}
\end{equation}
Upon completion of all agent processes, the AW yields the final result  $r$.
If $r$ matches the expected outcome, the AW is considered successful; otherwise, it is deemed unsuccessful.

\begin{definition}[Agentic Workflow Performance Prediction]
Given a specific task instruction $T$ and an AW $\mathcal{G}$, performance prediction aims to determine whether AW $\mathcal{G}$ can produce the expected outcome for task instruction $T$ without actually executing the AW. 
\end{definition}

The AW performance prediction provides a computationally efficient proxy that guides automatic AW generation while avoiding the substantial overhead of direct execution.

\section{Methodology}
\label{sec:method}
In this section, we introduce our proposed AW performance prediction method, GLOW. As illustrated in Fig. \ref{fig:architecture}, GLOW takes an AW and a task instruction as input and outputs a scalar performance score.  
Specifically, for AWs, high-level semantic representations are derived from a graph-oriented LLM, while structural dependencies are captured by a GNN. The representation of task instruction $T$ is extracted using a sentence-BERT. These distinct representations are then projected into a unified latent space and aggregated through a representation fusion module to generate the predicted performance score.
We next describe the representation encoding, performance prediction, and model training components in detail.

\subsection{Representation Encoding}
\label{sec:extraction}
GLOW encodes representations from the task instruction and the AW to support subsequent performance prediction.

\subsubsection{Task Instruction Encoding}  
Given the task instruction $T$, a pretrained Sentence-BERT (SBERT) \cite{reimers2019sentence} is first employed to obtain its semantic embedding, which captures the global task intent. To align this embedding with the latent space of the AW representations, a lightweight MLP is applied as the projector. The final task representation $\mathbf{R}^{\text{Task}} \in \mathbb{R}^{d}$,  where $d$ denotes the dimensionality of the representation, is formulated as: 
\begin{equation}
    \mathbf{R}^{\text{Task}} = \text{Proj}_{T}(\text{SBERT}(T))
\end{equation}
where $\text{Proj}_{T}(\cdot)$ denotes the projector.

\subsubsection{Agentic Workflow Structural Encoding} 
To capture structural dependencies and information propagation among nodes in the AW, a GNN is employed.
Initially, for each node $v_i$, its textual prompt $p_i$ is encoded by the Sentence-BERT to obtain the initial node embedding:
\begin{equation}
\mathbf{h}_{i}^{(0)} = \text{SBERT}(p_i)
\end{equation}
Subsequently, a GNN encodes the graph structure by propagating information along the edges $\mathcal{E}$. Specifically, the GNN performs $L$  layers of message passing over  $\mathcal{E}$, where node embeddings are iteratively updated by aggregating information from their upstream dependent nodes. After  $L$  layers, refined node embeddings for all nodes are obtained, formulated as:
\begin{equation}
    \{ \mathbf{h}_{i}^{(L)} \}_{v_i \in \mathcal{V}} = \text{GNN}(\{ \mathbf{h}_{i}^{(0)} \}_{v_i \in \mathcal{V}}, \mathcal{E})
\end{equation}
To derive the global structural representation $\mathbf{R}^{\text{GNN}} \in \mathbb{R}^{d}$, mean pooling \cite{xu2018powerful} is performed over all node embeddings, averaging them across all nodes.
\begin{equation}
     \mathbf{R}^{\text{GNN}} = \frac{1}{ |\mathcal{V}|} \sum_{v_i \in \mathcal{V}} \mathbf{h}_{i}^{(L)}
\end{equation}
where $|\mathcal{V}|$ denotes the total number of nodes in the AW.

\begin{figure}[t]
    \centering
    \includegraphics[scale=0.71]{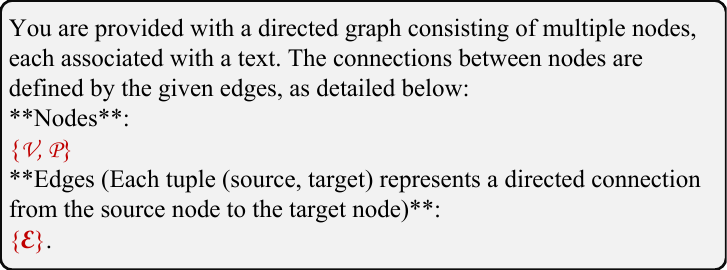}
    \caption{The template used to convert the AW into descriptive text. Node set $\mathcal{V}$ and prompt set $\mathcal{P}$ are organized into a dictionary mapping each node ID to its textual prompt, while the edge set ${\mathcal{E}}$ is converted into a list of (\textit{source}, \textit{target}) tuples.}
    \label{fig:prompt_template}
      \vskip -0.1in
\end{figure}

\subsubsection{Agentic Workflow Semantic Encoding}
While GNNs explicitly model structural dependencies, they primarily focus on local message passing and may not fully capture the high-level semantics embedded in node prompts and cross-node functional design. To complement this, the topology-aware semantic encoding capability of a graph-oriented LLM is leveraged.
Specifically, the AW $\mathcal{G}$ is linearized into a structured descriptive text $S_{\mathcal{G}}$, which comprises both node-level prompts and edge-level dependencies, following the template in Fig. \ref{fig:prompt_template}. 
To encourage the LLM to produce a compact graph-level embedding, a specific instruction is appended:
``\textit{Provide a single token representing the embedding of this graph.}''.
The resulting sequence is fed into the graph-oriented LLM, which has been instruction-tuned on graph understanding tasks to better capture topology-aware semantics, as described in Section \ref{sec:INLLM}.  
The final hidden state of the generated token, denoted as $\mathbf{h}^{\text{[EOS]}}$, is extracted and passed through a projector (implemented as an MLP) to obtain the semantic representation $\mathbf{R}^{\text{LLM}} \in \mathbb{R}^{d}$:
\begin{equation}
\mathbf{R}^{\text{LLM}} = \text{Proj}_{L}(\mathbf{h}^{\text{[EOS]}}),
\end{equation} 
This design follows prior studies \cite{jiang2024scaling,zhang2024simple} showing that the hidden states of generated tokens can serve as compact representations for downstream tasks.

\subsection{Performance Prediction}
\label{sec:prediction}
To synthesize the semantic and structural representations of the AW along with the task instruction representations, a Transformer-based fusion module is employed, followed by a prediction head that outputs the predicted performance score $\hat{y}$.

Specifically, a learnable prediction token representation $\mathbf{R}^{\text{Pred}}\in \mathbb{R}^{d}$ is introduced, serving as a global aggregator for all representations. It is stacked with the task instruction, AW structural, and AW semantic representations to form the initial input:
\begin{equation}
    \mathbf{R}^{(0)} = [\mathbf{R}^{\text{Pred}}; \mathbf{R}^{\text{LLM}}; \mathbf{R}^{\text{GNN}}; \mathbf{R}^{\text{Task}}]\in \mathbb{R}^{4\times d}
\end{equation}
To enable the model to distinguish between different types of representations, learnable type embeddings $\mathbf{E}^{\text{Type}} \in \mathbb{R}^{4 \times d}$ are added to $\mathbf{R}^{(0)}$.
The resulting representation sequence is then fed into a representation fusion module composed of $L_T$ stacked layers.
Each layer enables cross-representation interaction through Multi-Head Self-Attention (MHSA), allowing each component to attend to all others and dynamically exchange information, followed by a position-wise Feed-Forward Network (FFN). Both submodules are equipped with residual connections and Layer Normalization (LN). Formally, for the $l$-th layer, the representation update is given by: 
\begin{align}
    \tilde{\mathbf{R}}^{(l)} &= \text{LN}(\text{MHSA}(\mathbf{R}^{(l-1)}) + \mathbf{R}^{(l-1)}) \\
    \mathbf{R}^{(l)} &= \text{LN}(\text{FFN}(\tilde{\mathbf{R}}^{(l)}) + \tilde{\mathbf{R}}^{(l)})
\end{align}
Through this deep interaction, the prediction token aggregates context-aware information from all other representations. The final hidden state of the prediction token, denoted as $\mathbf{r}_{\text{Pred}}^{(L_T)}$, is fed into a Prediction Head (PH), implemented as an MLP,  followed by a sigmoid function to produce the final score $\hat{y}$: 
\begin{equation}
    \hat{y} =  \operatorname{Sigmoid}(\text{PH}(\mathbf{r}_{\text{Pred}}^{(L_T)}))
\end{equation}

\subsection{Model Training}
\label{sec:training}
To improve representation quality and cross-module alignment, we adopt a training strategy consisting of two independent pretraining procedures (i.e., LLM instruction-tuning and GNN pretraining) and one subsequent end-to-end model training stage. 
Specifically, the LLM instruction-tuning optimizes the LLM using graph understanding tasks to obtain a graph-oriented LLM, while the GNN pretraining independently optimizes the GNN using node and edge reconstruction objectives. 
After obtaining these pretrained components, the end-to-end model training jointly optimizes GLOW for the AW performance prediction. As shown in Fig. \ref{fig:architecture}, during this stage, SBERT and the graph-oriented LLM are frozen, while the GNN, projectors, representation fusion module, and prediction head are updated.

\subsubsection{LLM Instruction-Tuning}
\label{sec:INLLM}
To equip a generic LLM with a stronger ability to understand graph structures from plain text, we instruction-tune it using the descriptive text of AW  $S_{\mathcal{G}}$ generated from the template in Fig. \ref{fig:prompt_template}, and construct graph-related QA instances spanning six task types:
\begin{itemize}
   \item \textbf{Node Prompt Retrieval (NPR):} Retrieving the raw prompt of the specified node.
    \item \textbf{Directed Neighbor Extraction (DNE):} Identifying in-neighbors (predecessors) and out-neighbors (successors) for the specific node.
   \item \textbf{Subgraph Reachability \& Path Length (REACH):} Determining reachability between node pairs and predicting their shortest directed path length.
    \item \textbf{Degree-Based Prediction (DBP):} Predicting the node's in-degree, out-degree, and the graph's average degree.
   \item \textbf{Key Node Identification (KNI):} Identifying source nodes (zero in-degree) and sink nodes (zero out-degree).
   \item \textbf{Topological Sorting (TSORT):} Predicting a valid topological ordering of the nodes. 
\end{itemize}

These task types provide complementary supervision for learning topology-aware semantic representations. Specifically, NPR helps the  LLM establish the correspondence between node identifiers and node-level prompts, preserving the semantic information of individual agents. DNE and REACH provide supervision for modeling inter-agent dependencies and information propagation paths, enabling the model to capture how agents interact and exchange information within an AW. DBP and KNI focus on local and global structural characteristics, helping the model recognize important graph patterns. TSORT further enhances the understanding of global execution order by capturing the relative ordering constraints among agents. Although these task types partially overlap in terms of graph understanding capabilities, they provide complementary supervision from different perspectives, including node-level semantics, dependency relationships, and workflow-level structures. Together, these tasks encourage the  LLM to associate semantic information with graph structures and learn more informative topology-aware semantic representations for downstream AW performance prediction.

The LLM is instruction-tuned via Supervised Fine-Tuning (SFT) on the automatically constructed QA instances, where all answers are derived from deterministic graph rules, under the standard next-token prediction objective, resulting in a graph-oriented LLM.

\subsubsection{GNN Pretraining}
Before the end-to-end model training, the GNN is pretrained using self-supervised learning to ensure robust structural embeddings.
During this stage, SBERT is kept frozen, and the initial node embeddings  $\{ \mathbf{h}_{i}^{(0)} \}_{v_i \in \mathcal{V}}$ are generated by SBERT.
The objective includes two components: node reconstruction and edge reconstruction.

For \textbf{node reconstruction}, the initial node embeddings $\{ \mathbf{h}_{i}^{(0)} \}_{v_i \in \mathcal{V}}$ are reconstructed from the GNN output. Let $\mathbf{h}_i^{(L)}$ be the output embedding of node $v_i$ from the GNN. The Mean Squared Error (MSE) loss is minimized:
\begin{equation}
    \mathcal{L}_{\text{Node}} = \frac{1}{|\mathcal{V}|} \sum_{v_i \in \mathcal{V}} \| \text{Proj}(\mathbf{h}_i^{(L)}) - \mathbf{h}_i^{(0)} \|^2
\end{equation}
where $\text{Proj}(\cdot)$ is an auxiliary projection head, implemented as an MLP.

For \textbf{edge reconstruction}, a bilinear decoder is employed to predict the existence of directed edges. The probability of an edge from $v_i$ to $v_j$ is computed as:
\begin{equation}
    \hat{e}_{ij} = \operatorname{Sigmoid}(\mathbf{h}_i^{(L)\text{T}} \mathbf{W}\mathbf{h}_j^{(L)} + b )
\end{equation}
where $\mathbf{W}$ and  $b$ are the learnable weight matrix and bias, respectively, $\cdot^\text{T}$ represents transposition.
The Binary Cross-Entropy (BCE) loss is then optimized over all possible node pairs:
\begin{equation} 
    \mathcal{L}_{\text{Edge}} = - \frac{1}{|\mathcal{V}|^2} \sum_{v_i,v_j \in \mathcal{V}} [e_{ij} \log \hat{e}_{ij} + (1-e_{ij}) \log (1-\hat{e}_{ij})]
\end{equation}
where $e_{ij}=1$ if there is an edge from $v_i$ to $v_j$, 0 otherwise. 

Finally, the loss for pretraining GNN is defined as:
\begin{equation}
    \mathcal{L}_{\text{Pre}} = \mathcal{L}_{\text{Node}} + \mathcal{L}_{\text{Edge}}
\end{equation} 

\subsubsection{End-to-End Model Training}
In this stage, the parameters of SBERT and the graph-oriented LLM are frozen to preserve their pretrained knowledge.

To supervise the performance estimation, we adopt a \textbf{prediction loss} using BCE to supervise the performance estimation. 
Given the ground truth label $y \in \{0, 1\}$ (where 1 indicates the AW successfully completes the task) and the predicted performance score $\hat{y}$:
\begin{equation}
    \mathcal{L}_{\text{Pred}} = -\frac{1}{S} \sum_{i=1}^{S} [y_i \log \hat{y_i} + (1-y_i) \log (1-\hat{y_i})]
\end{equation}
where $S$ is the number of samples in the dataset.

Although successful AWs for the same task may exhibit diverse reasoning strategies, they often share latent structural and semantic regularities that correlate with task success \cite{cemri2026multi}. Encouraging these AWs to be closer in the latent space helps the model capture such regularities more effectively. 
In contrast, unsuccessful AWs arise from diverse failure modes (e.g., reasoning errors, missing dependencies, or suboptimal coordination patterns), resulting in highly heterogeneous representations in the latent space. Their relationships in the latent space are largely unconstrained and do not exhibit consistent structural regularities.
Therefore, to further refine the latent space, we apply contrastive learning to make the representations of successful AWs (i.e., those with $y=1$) cluster more tightly, while pushing them away from unsuccessful ones (i.e., those with  $y=0$). 
To achieve this, for each task instruction $T$, we construct a triplet set $\mathcal{T}_T$.  AWs with $y=1$ are selected as anchors. For each anchor, the positive sample $p$ is another successful AW with the same task, whereas the negative sample $n$ is an unsuccessful AW that fails to complete the task. The resulting \textbf{contrastive loss} is defined as:
\begin{equation}
\small
    \mathcal{L}_{\text{Con}}^{m} = \frac{1}{|\mathcal{T}_T|}  \sum_{(a, p, n) \in \mathcal{T}_T} 
    \max \left(0, d\left(\mathbf{R}_a^m, \mathbf{R}_p^m\right)-d\left(\mathbf{R}_a^m, \mathbf{R}_n^m\right)+\alpha\right)
\end{equation}
where $m \in \{\text{GNN}, \text{LLM}\}$ denotes the latent space of AW from either the GNN branch or the LLM branch, $d(\cdot,\cdot)$ represents a distance function (implemented as cosine distance), and $\alpha$ is a margin hyperparameter. 

The final training objective is defined as a weighted combination of the prediction loss and contrastive losses over the two AW representation branches:
\begin{equation}
\mathcal{L} = \mathcal{L}_{\text{Pred}} + \frac{\lambda}{2}( \mathcal{L}_{\text{Con}}^{\text{GNN}}+\mathcal{L}_{\text{Con}}^{\text{LLM}})
\end{equation}
where the hyperparameter $\lambda$ controls the trade-off between predictive accuracy and representation alignment.
During optimization, $\mathcal{L}_{\text{Con}}^{\text{GNN}}$ and $\mathcal{L}_{\text{Con}}^{\text{LLM}}$ respectively regularize the GNN branch and the LLM branch to learn more discriminative AW representations, while $\mathcal{L}_{\text{Pred}}$ supervises the entire model for the performance prediction.

\begin{table*}[t]\renewcommand{\arraystretch}{}
\caption{Statistics of FLORA-Bench used for performance prediction evaluation.}
\label{tab:stastic}
\setlength{\tabcolsep}{1mm}{
\begin{center}
\vskip -0.1in
\begin{tabular}{ccccccc}
\toprule
Domain            & Coding-GD & Coding-AF & Math-GD & Math-AF & Reason-GD & Reason-AF \\ \midrule
Num. of workflows & 1026      & 56        & 155     & 41      & 189       & 30        \\
Avg. of nodes     & 5.96      & 7.48      & 6.12    & 5.49    & 6.58      & 5.87      \\
Num. of tasks     & 57        & 233       & 97      & 99      & 2400      & 2400      \\
Num. of samples   & 30,683    & 7,362     & 12,561  & 4,059   & 453,600   & 72,000     \\  \bottomrule
\end{tabular}
\end{center}
} 
\vskip -0.1in
\end{table*}

\section{Experiments}
\label{sec:exp}
In this section, extensive experiments are conducted to investigate the following Research Questions (RQs):
\begin{itemize}
    \item \textbf{RQ1}: How effective is GLOW in predicting the performance of AWs?
    \item \textbf{RQ2}: How does instruction-tuning enhance the LLM’s capability to understand AWs from plain text?
    \item \textbf{RQ3}: How do different architectural components impact the overall performance of GLOW?
    \item \textbf{RQ4}: How do GNN pretraining and LLM instruction-tuning  contribute to the performance of GLOW?
    \item \textbf{RQ5}: How does incorporating contrastive loss affect the AW representations?
    \item \textbf{RQ6}: How do the hyperparameters $\alpha$ and $\lambda$ affect the performance of GLOW?
    \item \textbf{RQ7}: How effectively does GLOW support the downstream task, i.e., automatic AW generation?
\end{itemize}

GLOW is implemented in Python, and the source code is publicly available at https://github.com/guanwei49/GLOW.

\subsection{Experimental Setup} 
\subsubsection{Dataset} 
We adopt the recently introduced and well-curated FLORA-Bench benchmark \cite{zhang2025gnns}. It spans five representative datasets frequently studied in the agentic workflow literature, covering three core task types: code generation (HumanEval \cite{chen2021evaluating}, MBPP \cite{austin2021program}), mathematical problem solving (GSM8K \cite{cobbe2021training}, MATH \cite{hendrycks2021measuring}), and general reasoning (MMLU \cite{hendrycks2020measuring}). 
The AWs are derived from two state-of-the-art automatic AW generation methods: G-Designer (GD) \cite{zhang2024g} and AFLOW (AF) \cite{zhang2024aflow}. Table \ref{tab:stastic} summarizes the dataset statistics.
Each sub-dataset is randomly split into training, validation, and test sets following an 8:1:1 ratio.

In addition, to construct the dataset for \textbf{LLM instruction-tuning} (Section \ref{sec:INLLM}), all AWs from the FLORA-Bench are aggregated, resulting in 1,497 unique AWs. A random subset of 200 AWs is selected as the test set, while the remaining AWs are used as the training set.
For dataset generation, to ensure answer consistency and deterministic supervision, we explicitly specify in the instructions that node IDs should be returned in ascending order whenever multiple nodes appear in the answer.
For each AW, QA samples are instantiated according to the characteristics of each task type. Specifically, for DBP, DNE, NPR, and REACH, three distinct samples per task type are generated by varying the queried nodes or node pairs.
For KNI, two samples are generated by querying either source nodes or sink nodes.
For TSORT, each AW yields a single QA instance due to its global nature, where only one deterministic topological ordering is constructed under our lexicographical sorting strategy.
Overall, this process produces 19,455 instruction-tuning samples for training and 3,000 samples for testing.

\subsubsection{Baseline Methods}
We compare our method with the approach in \cite{zhang2025gnns}, where the GNN backbones include GCN \cite{kipf2016semi}, GAT \cite{velivckovic2017graph}, GCNII \cite{chen2020simple}, Graph Transformer (GT) \cite{shi2020masked}, and One-For-All (OFA) \cite{liu2023one}. We further compare against the Agentic Predictor (AP) \cite{trirat2025agentic} as an additional baseline.
In addition, we evaluate an LLM baseline using Qwen3-1.7B \cite{yang2025qwen3}\footnote{https://huggingface.co/Qwen/Qwen3-1.7B}. Specifically, we serialize both the AW and the task instruction into textual descriptions and directly feed them into the LLM. The LLM is then fine-tuned  to predict whether the given AW can successfully complete the task.  To reduce memory consumption during fine-tuning, we employ QLoRA \cite{dettmers2023qlora}, setting the LoRA rank to 8 and alpha to 32, with a batch size of 8.

\subsubsection{Implementation Details}
All experiments are conducted on a server equipped with an Intel Xeon Gold 6330 CPU (38 cores), 256GB of memory, and an NVIDIA A40 GPU with 48 GB of memory.
We utilize \textit{all-MiniLM-L6-v2}\footnote{https://huggingface.co/sentence-transformers/all-MiniLM-L6-v2} as the SBERT, Qwen3-1.7B as the base LLM and a two-layer GAT as the GNN. The representation fusion module consists of $L_T = 2$ stacked layers.
QLoRA \cite{dettmers2023qlora} is employed to reduce memory consumption during instruction-tuning, with LoRA rank set to 8 and LoRA alpha set to 32.
The hyperparameter $\lambda$, which balances the prediction loss and contrastive loss, is set to 1, while $\alpha$, controlling the margin in the contrastive loss, is set to 0.2.
The hidden dimension $d$ is 256, and the learning rate is $1 \times 10^{-4}$. 
We use the AdamW optimizer \cite{loshchilov2017decoupled} to train the model with a mini-batch size of 512. The maximum number of training epochs is 200, with early stopping applied if there is no improvement on the validation set for 30 consecutive epochs.
For fairness, the hyperparameters of all compared methods are set according to their original papers.
Each experiment is run five times, and the mean and standard deviation are reported.

\subsubsection{Metrics} 
The method’s performance is evaluated using two metrics.

First, \textbf{accuracy} measures the prediction correctness:
  \begin{equation}
  Accuracy = \frac{1}{S} \sum_{i=1}^{S} \mathbb{I}(\hat{y}_i=y_i)
  \end{equation}
where $S$ is the number of test samples and  $\mathbb{I}(\cdot)$ returns 1 if the condition holds and 0 otherwise. 

Second, \textbf{utility} assesses the consistency between the predicted and ground-truth rankings of AWs, emphasizing the method’s ability to distinguish the relative quality of different AWs. For each AW, the success rate is defined as the proportion of tasks it successfully completes. 
Let $\mathcal{H}_k$ and $\hat{\mathcal{H}}_k$ denote the sets of top-$k$ AWs selected based on the ground-truth and predicted success rates, respectively. The utility is defined as the mean overlap ratio averaged over various $k$:
\begin{equation}Utility = \frac{1}{K} \sum_{k =1}^{K} \frac{|\mathcal{H}_k \cap \hat{\mathcal{H}}_k|}{k}
\end{equation}
where $K$ is the total number of AWs in the test dataset.

\begin{table*}[ht]\renewcommand{\arraystretch}{}
\caption{Experimental results (\%) on the six domains in FLORA-Bench. Accuracy (Acc.) and utility (Uti.) are reported. The best-performing results are highlighted in bold.}
\label{tab:overall}
\vskip -0.05in
   \resizebox{\linewidth}{!}{
\begin{tabular}{ccccccccccccc}
\toprule
                  & \multicolumn{2}{c}{Coding-GD} & \multicolumn{2}{c}{Coding-AF} & \multicolumn{2}{c}{Math-GD} & \multicolumn{2}{c}{Math-AF} & \multicolumn{2}{c}{Reason-GD} & \multicolumn{2}{c}{Reason-AF} \\ \cmidrule(lr){2-3} \cmidrule(lr){4-5} \cmidrule(lr){6-7} \cmidrule(lr){8-9}  \cmidrule(lr){10-11}   \cmidrule(lr){12-13}  
& Acc. & Uti. & Acc. & Uti. & Acc. & Uti. & Acc. & Uti. & Acc. & Uti. & Acc. & Uti. \\ \midrule
GCN   & $82.1 {\scriptstyle \pm 0.2}$ & $74.3 {\scriptstyle \pm 0.8}$ & $82.7 {\scriptstyle \pm 0.1}$ & $71.3 {\scriptstyle \pm 0.9}$ & $59.8 {\scriptstyle \pm 1.1}$ & $60.1 {\scriptstyle \pm 1.3}$ & $79.8 {\scriptstyle \pm 0.1}$ & $72.9 {\scriptstyle \pm 0.5}$ & $71.6 {\scriptstyle \pm 0.2}$ & $62.0 {\scriptstyle \pm 0.7}$ & $85.1 {\scriptstyle \pm 0.1}$ & $86.6 {\scriptstyle \pm 0.8}$ \\
GAT   & $83.3 {\scriptstyle \pm 0.5}$ & $75.1 {\scriptstyle \pm 0.5}$ & $82.9 {\scriptstyle \pm 0.4}$ & $72.1 {\scriptstyle \pm 0.6}$ & $59.4 {\scriptstyle \pm 0.8}$ & $58.7 {\scriptstyle \pm 1.2}$ & $79.4 {\scriptstyle \pm 0.2}$ & $72.2 {\scriptstyle \pm 0.3}$ & $71.1 {\scriptstyle \pm 0.1}$ & $62.4 {\scriptstyle \pm 0.4}$ & $85.0 {\scriptstyle \pm 0.2}$ & $87.4 {\scriptstyle \pm 0.5}$ \\
GCNII & $82.4 {\scriptstyle \pm 0.3}$ & $75.4 {\scriptstyle \pm 0.7}$ & $82.2 {\scriptstyle \pm 0.2}$ & $71.6 {\scriptstyle \pm 0.8}$ & $61.0 {\scriptstyle \pm 0.7}$ & $59.1 {\scriptstyle \pm 0.9}$ & $78.4 {\scriptstyle \pm 0.1}$ & $72.5 {\scriptstyle \pm 0.6}$ & $71     .7 {\scriptstyle \pm 0.3}$ & $62.1 {\scriptstyle \pm 0.6}$ & $85.2 {\scriptstyle \pm 0.1}$ & $87.5 {\scriptstyle \pm 0.7}$ \\
GT    & $83.2 {\scriptstyle \pm 0.1}$ & $75.2 {\scriptstyle \pm 0.6}$ & $82.7 {\scriptstyle \pm 0.3}$ & $72.3 {\scriptstyle \pm 0.7}$ & $61.3 {\scriptstyle \pm 0.5}$ & $60.9 {\scriptstyle \pm 0.7}$ & $79.4 {\scriptstyle \pm 0.3}$ & $71.4 {\scriptstyle \pm 0.4}$ & $71.6 {\scriptstyle \pm 0.1}$ & $62.7 {\scriptstyle \pm 0.5}$ & $85.1 {\scriptstyle \pm 0.1}$ & $86.9 {\scriptstyle \pm 0.6}$ \\
OFA   & $82.3 {\scriptstyle \pm 0.4}$ & $74.1 {\scriptstyle \pm 0.4}$ & $82.2 {\scriptstyle \pm 0.5}$ & $72.8 {\scriptstyle \pm 0.5}$ & $60.0 {\scriptstyle \pm 0.6}$ & $59.9 {\scriptstyle \pm 0.8}$ & $78.9 {\scriptstyle \pm 0.1}$ & $69.8 {\scriptstyle \pm 0.5}$ & $70.9 {\scriptstyle \pm 0.2}$ & $62.7 {\scriptstyle \pm 0.3}$ & $84.3 {\scriptstyle \pm 0.3}$ & $86.3 {\scriptstyle \pm 0.4}$ \\
Qwen3 & $84.2 {\scriptstyle \pm 0.2}$ & $76.1 {\scriptstyle \pm 0.9}$ & $81.4 {\scriptstyle \pm 0.1}$ & $72.4 {\scriptstyle \pm 1.0}$ & $62.0 {\scriptstyle \pm 0.3}$ & $61.4 {\scriptstyle \pm 0.4}$ & $76.7 {\scriptstyle \pm 0.2}$ & $70.4 {\scriptstyle \pm 0.5}$ & $71.8 {\scriptstyle \pm 0.1}$ & $62.6 {\scriptstyle \pm 0.4}$ & $84.1 {\scriptstyle \pm 0.1}$ & $88.7 {\scriptstyle \pm 0.9}$ \\
AP & $83.4 {\scriptstyle \pm 0.2}$ & $75.9 {\scriptstyle \pm 0.7}$ & $83.2 {\scriptstyle \pm 0.2}$ & $73.9 {\scriptstyle \pm 0.8}$ & $62.9 {\scriptstyle \pm 0.4}$ & $61.8 {\scriptstyle \pm 0.3}$ & $79.8 {\scriptstyle \pm 0.2}$ & $73.4 {\scriptstyle \pm 0.4}$ & $72.6 {\scriptstyle \pm 0.2}$ & $63.1 {\scriptstyle \pm 0.5}$ & $85.7 {\scriptstyle \pm 0.1}$ & $88.1 {\scriptstyle \pm 0.7}$ \\
GLOW  & $\textbf{85.1} {\scriptstyle \pm 0.3}$ & $\textbf{77.3} {\scriptstyle \pm 0.6}$ & $\textbf{84.6} {\scriptstyle \pm 0.3}$ & $\textbf{75.4} {\scriptstyle \pm 0.7}$ & $\textbf{64.4} {\scriptstyle \pm 0.2}$ & $\textbf{63.5} {\scriptstyle \pm 0.5}$ & $\textbf{81.3} {\scriptstyle \pm 0.1}$ & $\textbf{75.1} {\scriptstyle \pm 0.4}$ & $\textbf{73.8} {\scriptstyle \pm 0.1}$ & $\textbf{66.1} {\scriptstyle \pm 0.5}$ & $\textbf{87.0} {\scriptstyle \pm 0.1}$ & $\textbf{90.5} {\scriptstyle \pm 0.6}$ \\ \bottomrule
\end{tabular}}
\end{table*}

\subsection{Performance Evaluation (RQ1)} 
The quantitative results are summarized in Table \ref{tab:overall}. Overall, GLOW consistently outperforms all baseline methods across all domains and evaluation metrics. 
Compared with the strongest baseline, AP, GLOW achieves average improvements of about 1.5\% in accuracy and 2.0\% in utility, demonstrating its effectiveness in identifying high-quality AWs and its suitability as a proxy model for execution-based evaluation in downstream automatic AW generation.
In comparison, traditional GNN-based methods (e.g., GCN, GAT) and AP achieve reasonable performance but are limited in capturing rich semantic information in agent prompts. While Transformer-based GT provides some improvements, it still lacks explicit alignment between workflow structure and agent semantics. On the other hand, the LLM baseline (Qwen3) performs well in semantic understanding but struggles to model explicit graph structures, leading to suboptimal results on structure-sensitive settings.
Overall, these results demonstrate that GLOW effectively bridges structural modeling and topology-aware semantic encoding, leading to consistent improvements over existing baselines.

\begin{table*}[t]
\caption{Efficiency comparison in terms of  training and inference time and GPU memory usage.}
\label{tab:exp_eff}
\begin{center}
\vskip -0.1in
\renewcommand{\arraystretch}{1}
\setlength{\tabcolsep}{2.2mm}{
\begin{tabular}{ccccc}
\toprule
    & \multicolumn{2}{c}{Training} & \multicolumn{2}{c}{Inference} \\ 
        \cmidrule(lr){2-3}     \cmidrule(lr){4-5}
    & Time (s/Epoch)               & GPU Memory (GB)
            & Latency  (ms/Sample)             & GPU Memory (GB) 
          \\ \midrule
GCN      & 5.53  &   1.70  &   3.24  & 0.15 \\
GAT      & 5.98  &  1.85  &   3.47 & 0.16 \\
GCNII    & 5.47  &  1.87 &  3.09  &   0.16 \\
GT       & 6.62  &  2.76  &   4.31 &  0.24 \\
OFA      & 8.46  & 1.95  &   4.78 &  0.18 \\
Qwen3  &  1951.60  &  19.87 &  54.21  & 6.42 \\
AP       & 24.13  &  11.24 &   9.14 & 0.93  \\
GLOW     & 19.76 &  10.35 &  40.55  & 4.93\\
 \bottomrule
\end{tabular}}
\end{center}
\vskip -0.1in
\end{table*}

\subsection{Efficiency Comparison}
Table \ref{tab:exp_eff} reports the average training and inference efficiency across all domains in FLORA-Bench in terms of time and GPU memory usage. During training, we set the batch size of Qwen3-1.7B to 8 due to its significantly higher memory consumption, while all other methods adopt a batch size of 512. For inference, all methods are evaluated in a consistent setting where each sample is processed individually.
Overall, although GLOW incorporates an LLM, its training overhead remains moderate. This is mainly because the LLM is frozen during training, and each AW only requires a single forward pass to obtain its representation, without backpropagating through the LLM. In addition, representations of identical AWs are cached and reused, avoiding redundant forward computations and further improving training efficiency.
As a result, the computational cost of GLOW is dominated by lightweight modules (e.g., GNN and representation fusion module), making its training time comparable to GNN-based baselines and significantly lower than Qwen3.
During inference, GLOW achieves lower latency and memory usage than Qwen3. This improvement is mainly due to the shorter input sequences used in GLOW, where the LLM only encodes the descriptive text of AWs. In contrast, Qwen3 jointly processes both the descriptive text of  AW and the task instruction, leading to longer token sequences and consequently higher computational and memory overhead.
Although GLOW introduces higher latency and memory overhead compared to lightweight GNN-only baselines due to LLM-based representation encoding, this overhead is acceptable since AW performance prediction does not involve large-scale online requests and each prediction only requires less than 50ms, while achieving significantly superior performance.

\begin{table*}[t]\renewcommand{\arraystretch}{}
\caption{Experimental results (Accuracy, \%)  illustrating that the graph-oriented LLM, fine-tuned from the base LLM, achieves enhanced comprehension of AWs.}
\label{tab:graphLLM}
\setlength{\tabcolsep}{1mm}{
\begin{center}
\vskip -0.1in
\begin{footnotesize}
\begin{tabular}{cccccccc}\toprule
                  & NPR    & DNE     & REACH & DBP & KNI   & TSORT & Average \\ \midrule
Base LLM       & 36.3   & 93.7  & 93.2  & 65.3 & 85.3 & 21.5 & 65.9   \\
Graph-oriented LLM   & 100.0    & 100.0     & 98.7 & 97.0 & 99.7 & 99.0  & 99.1  \\ \bottomrule
\end{tabular}
\end{footnotesize}
\end{center}
}
\vskip -0.1in
\end{table*}

\begin{table*}[t]\renewcommand{\arraystretch}{}
\caption{Ablation results (\%), where `w/o' denotes removal of a component, and `w/o P.' indicates no pretraining or instruction-tuning.}
\label{tab:ab}
\vskip -0.05in
   \resizebox{\linewidth}{!}{
\begin{tabular}{ccccccccccccc}
\toprule
                  & \multicolumn{2}{c}{Coding-GD} & \multicolumn{2}{c}{Coding-AF} & \multicolumn{2}{c}{Math-GD} & \multicolumn{2}{c}{Math-AF} & \multicolumn{2}{c}{Reason-GD} & \multicolumn{2}{c}{Reason-AF} \\ \cmidrule(lr){2-3} \cmidrule(lr){4-5} \cmidrule(lr){6-7} \cmidrule(lr){8-9}  \cmidrule(lr){10-11}   \cmidrule(lr){12-13}  
& Acc. & Uti. & Acc. & Uti. & Acc. & Uti. & Acc. & Uti. & Acc. & Uti. & Acc. & Uti. \\ \midrule
w/o $\mathbf{R}^{\text{GNN}}$       & $83.8 {\scriptstyle \pm 0.2}$ & $76.0 {\scriptstyle \pm 0.7}$ & $82.4 {\scriptstyle \pm 0.4}$ & $73.2 {\scriptstyle \pm 0.8}$ & $62.4 {\scriptstyle \pm 0.1}$ & $61.4 {\scriptstyle \pm 0.6}$ & $77.4 {\scriptstyle \pm 0.1}$ & $72.1 {\scriptstyle \pm 0.5}$ & $72.0 {\scriptstyle \pm 0.2}$ & $63.2 {\scriptstyle \pm 0.4}$ & $85.0 {\scriptstyle \pm 0.1}$ & $87.6 {\scriptstyle \pm 0.5}$ \\
w/o $\mathbf{R}^{\text{LLM}}$       & $83.7 {\scriptstyle \pm 0.4}$ & $75.8 {\scriptstyle \pm 0.5}$ & $82.9 {\scriptstyle \pm 0.2}$ & $73.4 {\scriptstyle \pm 0.9}$ & $63.5 {\scriptstyle \pm 0.3}$ & $61.9 {\scriptstyle \pm 0.4}$ & $80.9 {\scriptstyle \pm 0.2}$ & $73.2 {\scriptstyle \pm 0.3}$ & $72.1 {\scriptstyle \pm 0.1}$ & $63.8 {\scriptstyle \pm 0.7}$ & $85.9 {\scriptstyle \pm 0.2}$ & $87.7 {\scriptstyle \pm 0.8}$ \\
w/o $\mathbf{E}^{\text{Type}}$            & $82.7 {\scriptstyle \pm 0.3}$ & $75.1 {\scriptstyle \pm 0.8}$ & $83.8 {\scriptstyle \pm 0.3}$ & $74.6 {\scriptstyle \pm 0.6}$ & $62.6 {\scriptstyle \pm 0.2}$ & $61.9 {\scriptstyle \pm 0.5}$ & $79.8 {\scriptstyle \pm 0.1}$ & $72.2 {\scriptstyle \pm 0.4}$ & $71.7 {\scriptstyle \pm 0.2}$ & $62.4 {\scriptstyle \pm 0.5}$ & $85.5 {\scriptstyle \pm 0.1}$ & $86.9 {\scriptstyle \pm 0.7}$ \\
w/o P. GNN        & $84.7 {\scriptstyle \pm 0.2}$ & $76.8 {\scriptstyle \pm 0.6}$ & $83.4 {\scriptstyle \pm 0.4}$ & $74.5 {\scriptstyle \pm 0.7}$ & $64.0 {\scriptstyle \pm 0.1}$ & $62.4 {\scriptstyle \pm 0.6}$ & $80.9 {\scriptstyle \pm 0.2}$ & $74.2 {\scriptstyle \pm 0.3}$ & $73.1 {\scriptstyle \pm 0.1}$ & $64.4 {\scriptstyle \pm 0.4}$ & $86.4 {\scriptstyle \pm 0.1}$ & $89.4 {\scriptstyle \pm 0.6}$ \\
w/o P. LLM        & $83.9 {\scriptstyle \pm 0.4}$ & $76.2 {\scriptstyle \pm 0.5}$ & $82.9 {\scriptstyle \pm 0.3}$ & $73.6 {\scriptstyle \pm 0.8}$ & $63.0 {\scriptstyle \pm 0.3}$ & $62.0 {\scriptstyle \pm 0.4}$ & $80.0 {\scriptstyle \pm 0.1}$ & $73.8 {\scriptstyle \pm 0.5}$ & $72.4 {\scriptstyle \pm 0.2}$ & $63.7 {\scriptstyle \pm 0.6}$ & $85.8 {\scriptstyle \pm 0.1}$ & $88.6 {\scriptstyle \pm 0.5}$ \\
w/o P. GNN \& LLM & $83.7 {\scriptstyle \pm 0.3}$ & $75.9 {\scriptstyle \pm 0.7}$ & $82.7 {\scriptstyle \pm 0.2}$ & $73.1 {\scriptstyle \pm 0.6}$ & $62.7 {\scriptstyle \pm 0.2}$ & $61.7 {\scriptstyle \pm 0.5}$ & $79.8 {\scriptstyle \pm 0.1}$ & $73.1 {\scriptstyle \pm 0.4}$ & $72.4 {\scriptstyle \pm 0.1}$ & $63.6 {\scriptstyle \pm 0.5}$ & $85.4 {\scriptstyle \pm 0.1}$ & $88.1 {\scriptstyle \pm 0.6}$ \\
GLOW  & $\textbf{85.1} {\scriptstyle \pm 0.3}$ & $\textbf{77.3} {\scriptstyle \pm 0.6}$ & $\textbf{84.6} {\scriptstyle \pm 0.3}$ & $\textbf{75.4} {\scriptstyle \pm 0.7}$ & $\textbf{64.4} {\scriptstyle \pm 0.2}$ & $\textbf{63.5} {\scriptstyle \pm 0.5}$ & $\textbf{81.3} {\scriptstyle \pm 0.1}$ & $\textbf{75.1} {\scriptstyle \pm 0.4}$ & $\textbf{73.8} {\scriptstyle \pm 0.1}$ & $\textbf{66.1} {\scriptstyle \pm 0.5}$ & $\textbf{87.0} {\scriptstyle \pm 0.1}$ & $\textbf{90.5} {\scriptstyle \pm 0.6}$ \\ \bottomrule
\end{tabular}}
\vskip -0.1in
\end{table*}

\subsection{Impact of LLM Instruction-Tuning  (RQ2)}
To answer RQ2, the zero-shot performance of the vanilla base LLM is compared against the fine-tuned graph-oriented LLM. The results are reported in Table \ref{tab:graphLLM}.
The graph-oriented LLM achieves a near-perfect average accuracy of 99.1\%, significantly outperforming the base LLM (65.9\%). 
The performance gap is particularly evident in task types requiring precise ID-text mapping (i.e., NPR) and global structural understanding (i.e., TSORT). 
The base LLM performs poorly on NPR, achieving only 36.3\% accuracy, which indicates difficulty in aligning node IDs with their corresponding textual prompts. 
In contrast, the graph-oriented LLM reaches a perfect accuracy of 100\%, demonstrating that it has successfully learned the alignment between node IDs and their semantic content. 
A similar trend is observed in the TSORT, where the base LLM again underperforms with 21.5\% accuracy, whereas our graph-oriented LLM significantly improves to  99.0\%, indicating a strong ability to infer the global execution flow of the AW.
These results show that small-version LLMs (Qwen3-1.7B), despite strong linguistic reasoning, cannot inherently parse serialized graphs or capture topological dependencies without adaptation.
By adapting the LLM into a graph-oriented expert, the semantic representations fed to the downstream predictor are ensured to be not mere textual embeddings, but deeply grounded in the AW topology and interactions among agents.

\subsection{Ablation Studies}
\subsubsection{Architectural Component (RQ3)}
To assess the contribution of each architectural component, GLOW is compared with variants where specific components are removed. As shown in rows 1–3 of  Table \ref{tab:ab}, removing any component leads to a performance degradation.
The removal of  $\mathbf{R}^{\text{GNN}}$ results in the most significant impact, causing an average drop of 2.2\% in accuracy and 2.4\% in utility across the six domains. Similarly, excluding $\mathbf{R}^{\text{LLM}}$ results in an average decline of 1.2\% in accuracy and 2.0\% in utility. This quantitative evidence suggests that while structural information plays a dominant role, combining structural representations with topology-aware semantic representations further improves AW performance, thereby demonstrating the effectiveness of the proposed dual-branch representation encoding framework.
Furthermore, the absence of type embeddings (w/o $\mathbf{E}^{\text{Type}}$) results in an average decline of 1.2\% in accuracy and 2.0\% in utility. This result indicates that explicitly distinguishing representation types through learnable embeddings greatly improves the fusion module’s ability to integrate heterogeneous information.

\begin{figure*}[t]
    	\centering
    \subfloat[GLOW]{
	\centering
	\includegraphics[width=0.37\textwidth]{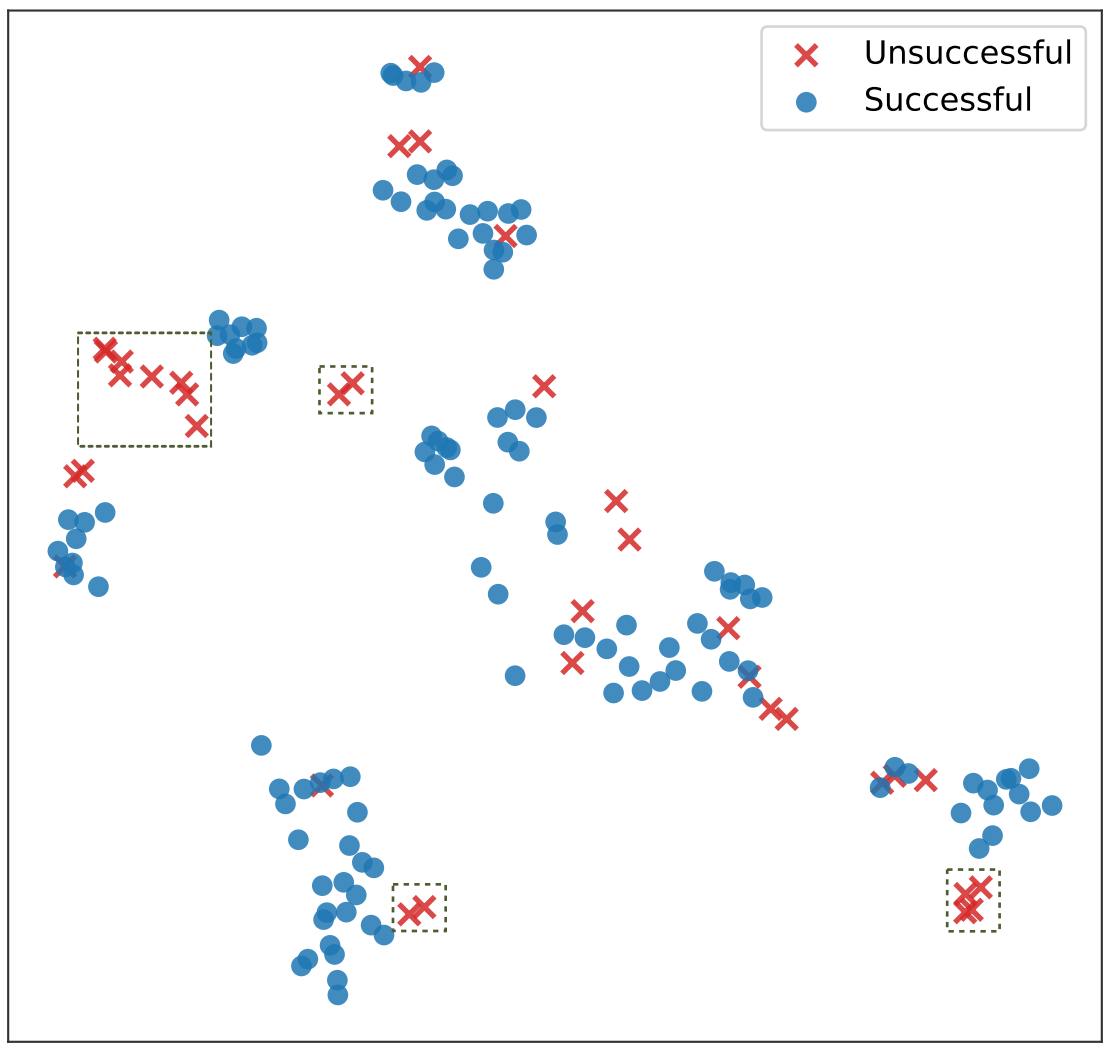}
	\label{fig:rep_mathgd}
	}
	\hfil
    \subfloat[Without contrastive loss]{
	\centering
	\includegraphics[width=0.37\textwidth]{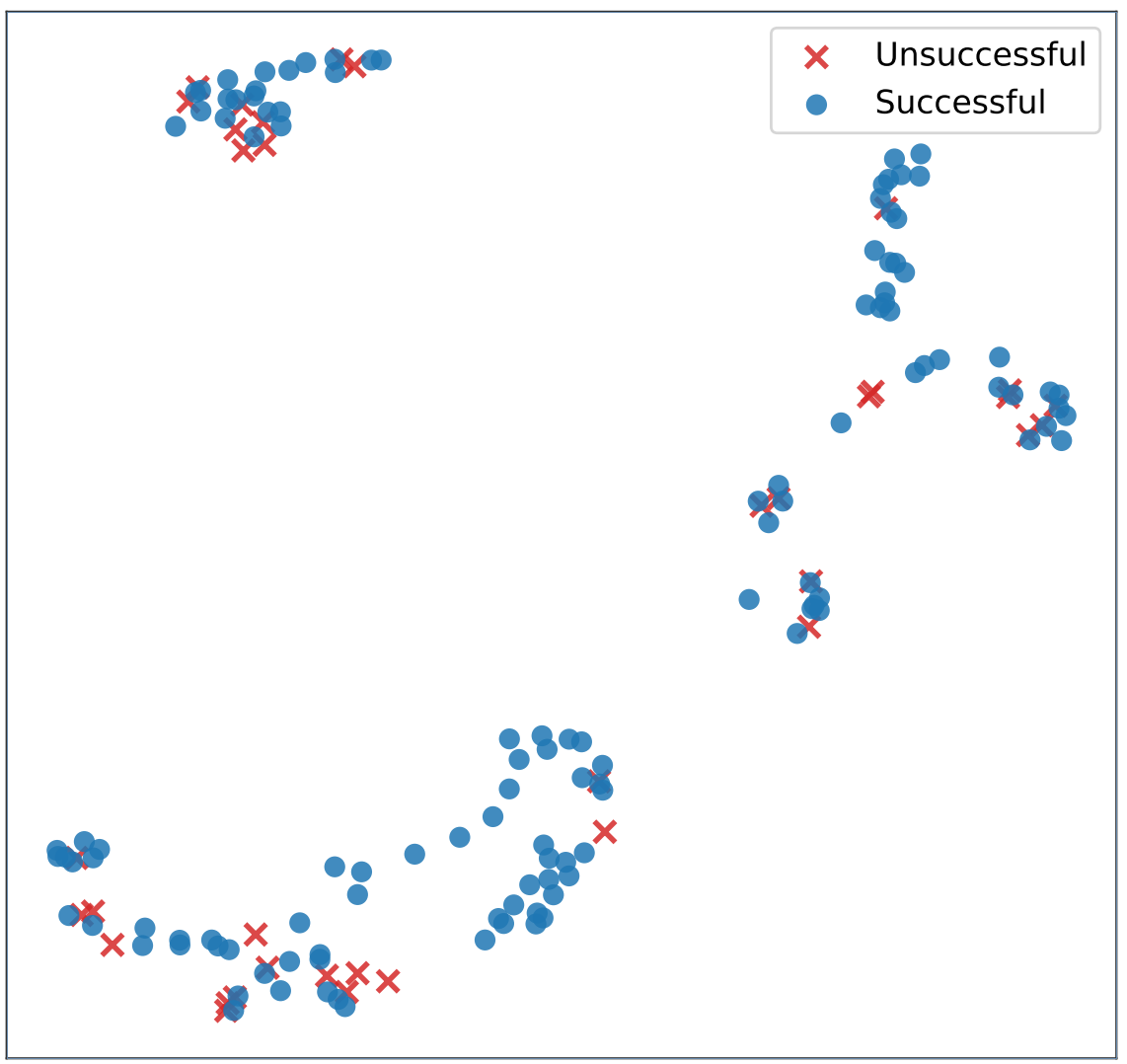}
	\label{fig:rep_mathgd-wo_con}
	}\\
	\vskip -0.05in
	\caption{Impact of contrastive loss on AW representations in Math-GD.  The representations are visualized using UMAP.}
 \label{fig:represen}
	\vskip -0.1in
\end{figure*}

\begin{figure*}[t]
    	\centering
    \subfloat{
	\centering
	\includegraphics[scale=0.4]{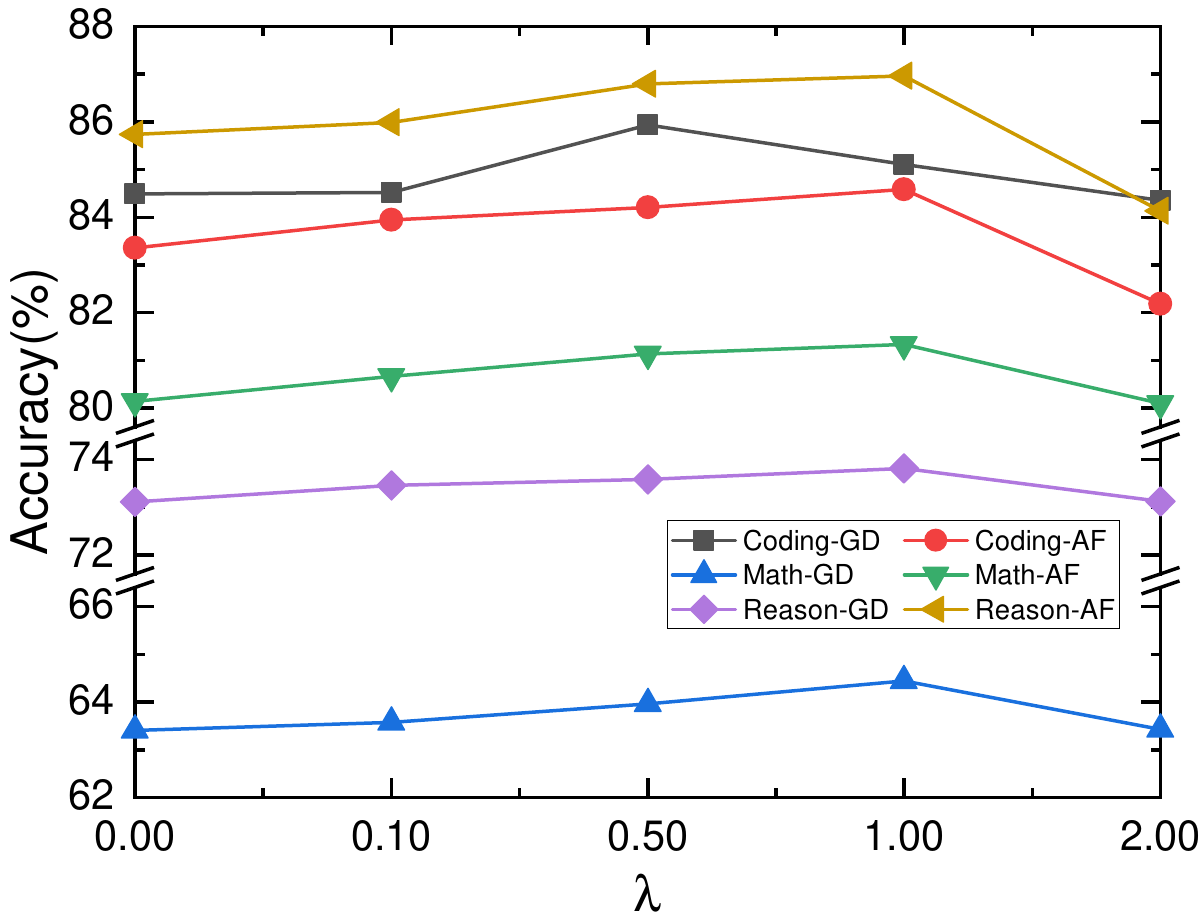}
	\label{fig:lossweightG}
	}
	\hfil
\subfloat{
	\centering
	\includegraphics[scale=0.4]{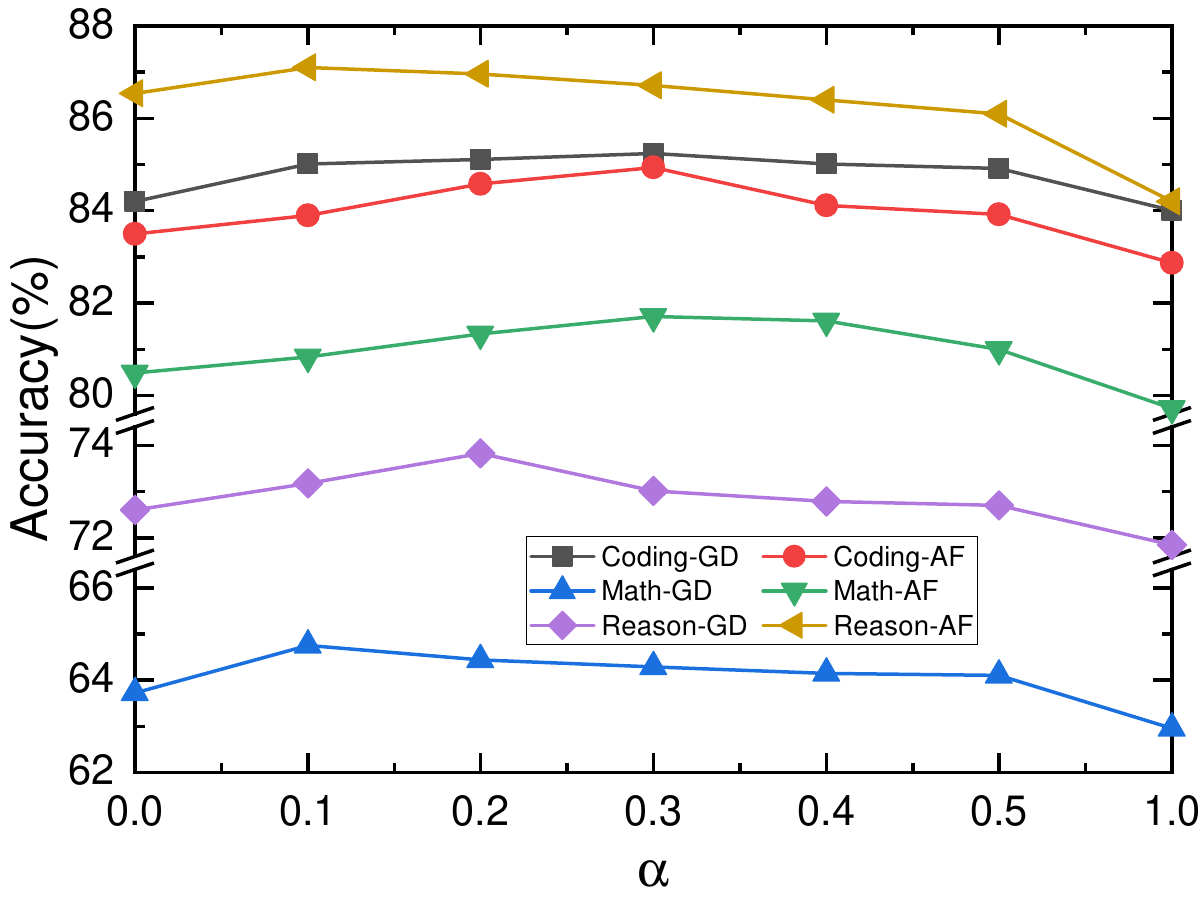}
	\label{fig:marginG}
	}\\
	\vskip -0.05in
	\caption{Impact of hyperparameters $\lambda$ and $\alpha$ on model performance.}
 \label{fig:hyper}
	\vskip -0.1in
\end{figure*}

\subsubsection{GNN Pretraining and LLM Instruction-Tuning (RQ4)}
As shown in rows 4–6 of  Table \ref{tab:ab}, the variant without LLM instruction-tuning (w/o P. LLM) and the variant without GNN pretraining (w/o P. GNN) consistently lead to performance degradation across all six domains, highlighting the importance of both pretraining strategies.
In particular, removing both GNN pretraining and LLM instruction-tuning (w/o P. GNN \& LLM) leads to an average drop of 1.6\% in accuracy and 2.1\% in utility across the six domains.
These results indicate that GNN pretraining enhances the GNN’s ability to capture structural dependencies, while LLM instruction-tuning improves its capacity for graph-aware semantic understanding. Moreover, the complementary nature of the two components suggests that jointly leveraging structural pretraining and instruction-level supervision is crucial for achieving robust performance in AW representation encoding.


\subsubsection{Impact of Contrastive Loss on AW Representations (RQ5)}
Fig.~\ref{fig:represen} illustrates the two-dimensional UMAP \cite{mcinnes2018umap} visualization of AW representations in Math-GD produced by GLOW and its ablation variant without contrastive loss, where AWs are labeled as successful or unsuccessful for a specific task. 
The representations learned without contrastive loss exhibit substantial overlap between successful and unsuccessful AWs, indicating limited discriminative capability. 
In contrast, GLOW produces more distinguishable representations, where unsuccessful AWs are more likely to lie outside the dominant successful clusters. Moreover, unsuccessful AWs also tend to form separate semantic clusters, as highlighted by the green dashed box. These results demonstrate that contrastive loss enhances the separability and structural organization of AW representations.

\subsection{Hyperparameter Study (RQ6)}
We examine GLOW’s sensitivity to two key hyperparameters: the loss weight $\lambda$, which balances the prediction and  contrastive losses, and the margin $\alpha$, which controls the contrastive separation. Fig. \ref{fig:hyper} reports the accuracy under different settings. Notably, \textbf{setting $\lambda = 0$ corresponds to the ablation of the contrastive loss}.
As expected, the accuracy exhibits a consistent trend with respect to both hyperparameters, first improving and then declining as the hyperparameter values become extreme.
Specifically, introducing the contrastive loss consistently improves performance compared with the setting $\lambda = 0$, with the best performance achieved when $\lambda \in [0.5, 1.0]$.
In addition, the model performs best when the margin is set within $\alpha \in [0.2, 0.3]$, since a small margin yields insufficient discrimination between positive and negative pairs, whereas an excessively large margin introduces overly restrictive constraints that hinder optimization and distort the latent space.
Importantly, the accuracy variation within these ranges is small, indicating that GLOW is robust and not overly sensitive to precise hyperparameter choices.
These results suggest that $\lambda = 1.0$ and $\alpha = 0.2$ provide reliable performance and are therefore adopted as the default settings.

\begin{figure*}[ht]
    	\centering
    \subfloat[HumanEval]{
	\centering
	\includegraphics[scale=0.27]{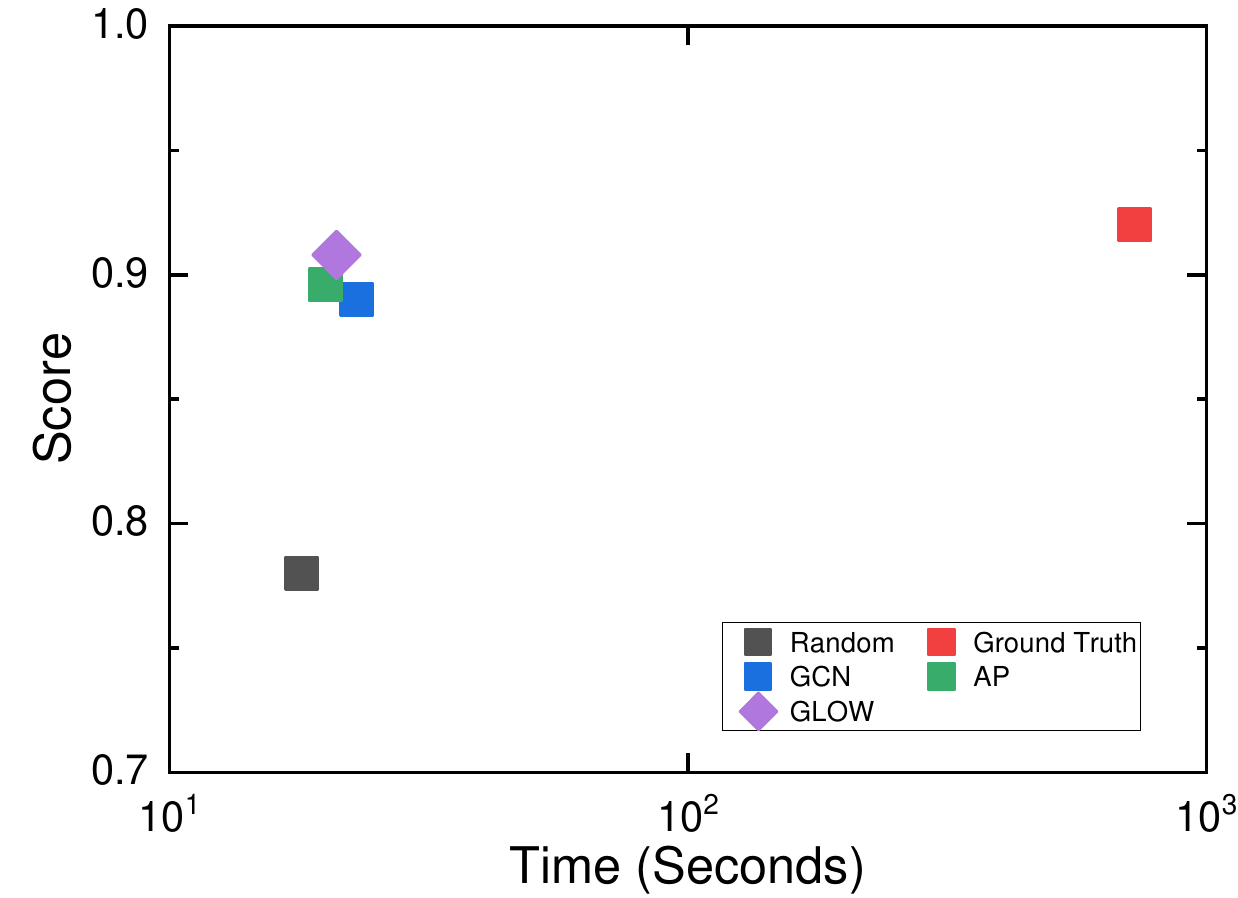}
	\label{fig:HumanEvalG}
	}
	\hfil
\subfloat[MBPP]{
	\centering
	\includegraphics[scale=0.27]{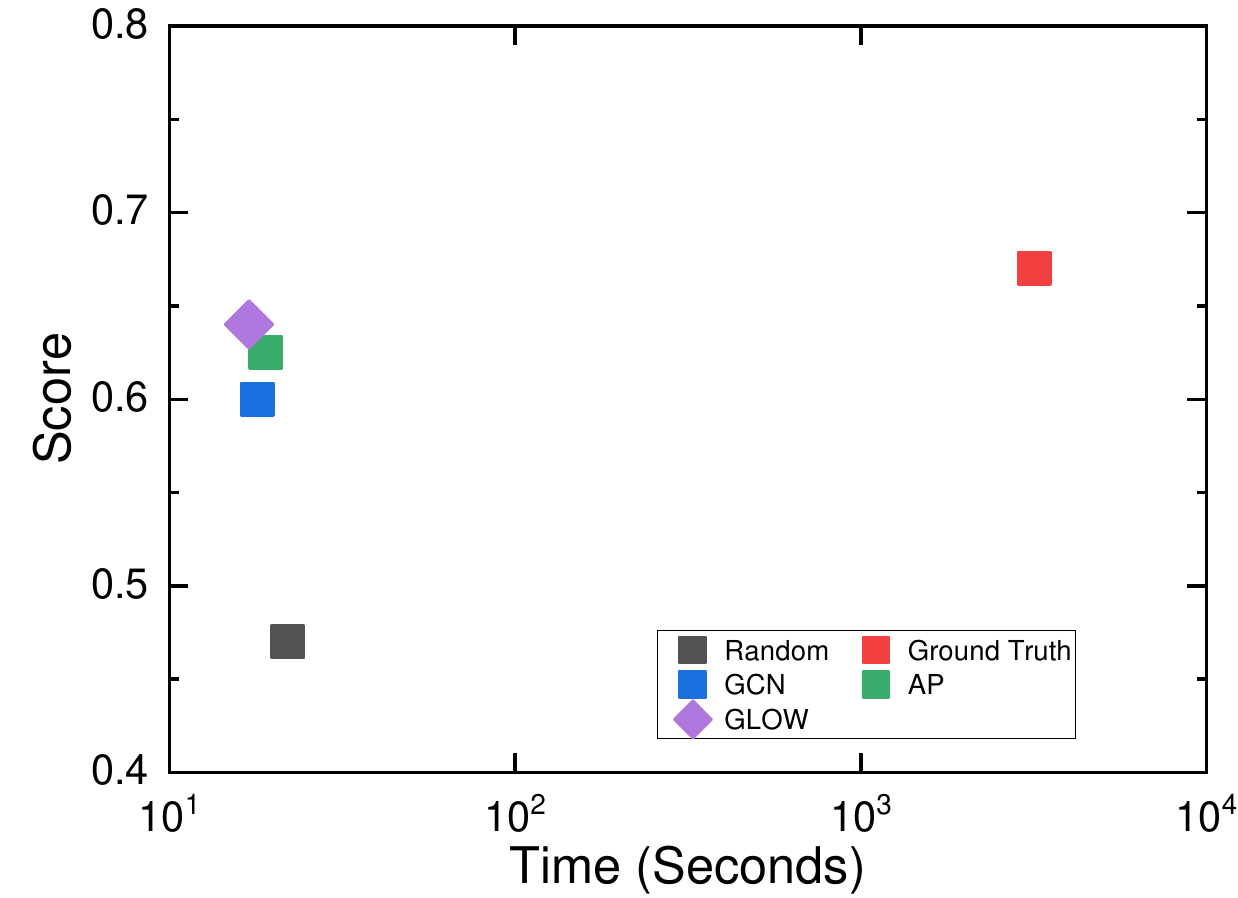}
	\label{fig:MBPPG}
	}
    	\hfil
\subfloat[MMLU]{
	\centering
	\includegraphics[scale=0.27]{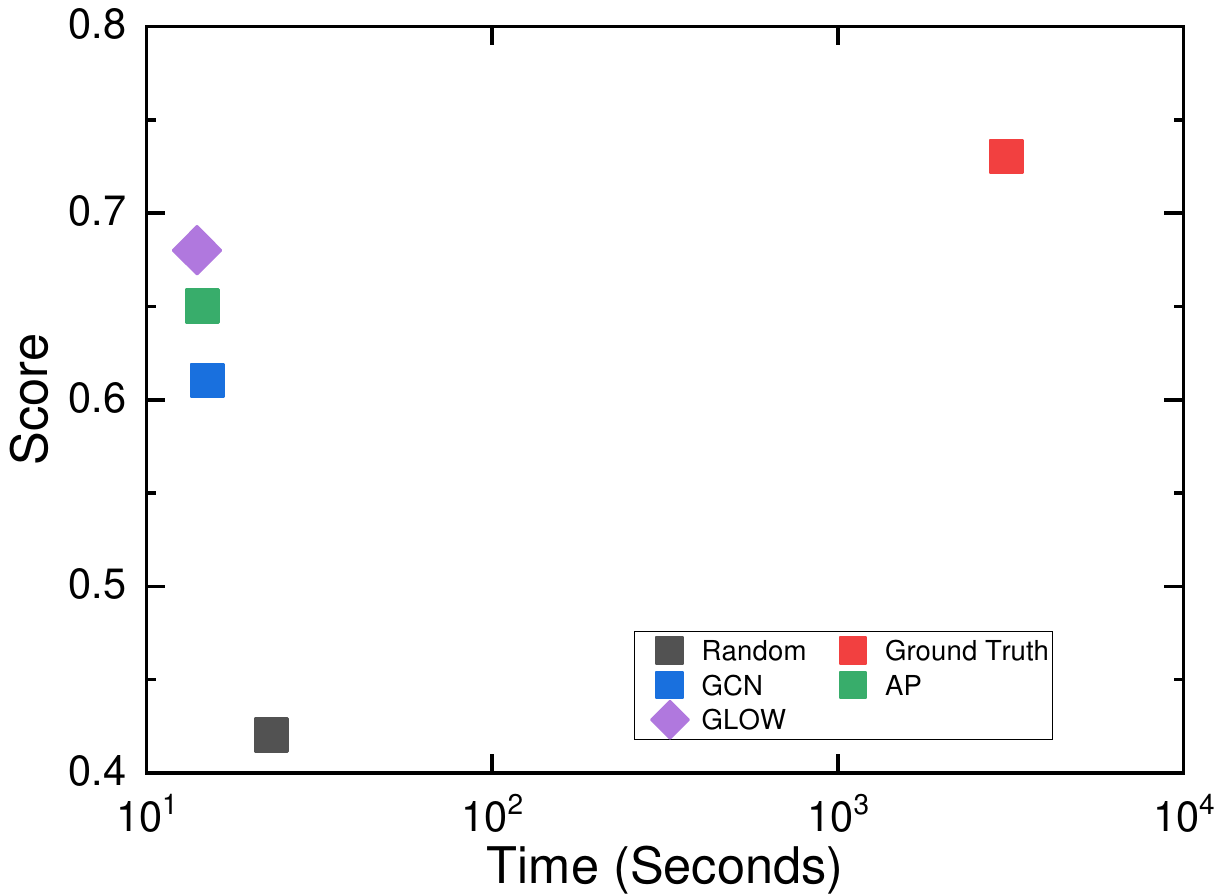}
	\label{fig:MMLUG}
	}
    \\
	\vskip -0.05in
	\caption{Comparison of time consumption and final AW performance across different AW evaluation methods in AFLOW.}
 \label{fig:pf}
	\vskip -0.1in
\end{figure*}

\subsection{Impact on Automatic AW Generation (RQ7)}
We evaluate the impact of different AW performance prediction methods on the automatic AW generation framework, AFLOW.
We compare GLOW against three baselines:
i) `Random', which predicts an AW’s performance uniformly at random;
ii) the standard `GCN'-based predictor; and
ii) the `Agentic Predictor' (AP).
We also include `Ground Truth', i.e., execution-based evaluation, which obtains the actual performance by executing the AW, to represent the upper bound of performance prediction methods for AFLOW.
The reported `Score' metric reflects the success rate of the final AWs generated by AFLOW on the test dataset.

As shown in Fig. \ref{fig:pf}, all performance prediction methods can effectively accelerate automatic AW generation, reducing runtime by more than 98\% by decreasing reliance on costly execution-based evaluation.  
From a performance perspective, GLOW consistently outperforms the Random, GCN, and AP baselines, owing to its more accurate performance predictions.
GLOW’s performance closely approaches the ceiling established by the Ground Truth, demonstrating that GLOW can effectively guide AFLOW toward high-quality AWs with minimal performance loss.
Moreover, compared with the computationally expensive Ground Truth, which requires repeated LLM calls, GLOW substantially accelerates AFLOW’s optimization process, reducing time consumption by 98.7\% while incurring only a 0.031 decrease in score on average across three datasets.
Notably, this performance degradation is significantly smaller than that of other performance prediction methods, where Random, GCN, and AP incur average score decreases of 0.217, 0.073, and 0.050, respectively.
Compared with the Random, GCN and AP, GLOW’s more reliable performance prediction also helps AFLOW converge slightly faster, as observed on datasets such as MBPP and MMLU.

Overall, these results show that although all performance prediction methods can speed up AFLOW, GLOW achieves a better balance between efficiency and effectiveness, with significantly smaller performance degradation. This demonstrates its practical value as a reliable proxy for accelerating automatic AW generation.

\section{Conclusion and Future Work}
\label{sec:conclusions}
In this paper, we propose GLOW, a unified framework for AW performance prediction that integrates the structural modeling ability of GNNs with the topology-aware semantic encoding capability of LLMs. By introducing a graph-oriented LLM through instruction-tuning, GLOW enhances the LLM’s ability to understand graph structures and capture agent-level semantic logic from textual workflow descriptions. 
Based on this, a dual-branch AW representation encoding module is further designed, with a GNN branch encoding structural dependencies and an LLM branch capturing semantic information, enabling joint modeling of structural and semantic aspects of AWs.
A contrastive learning strategy further enhances the discriminative quality of the learned representations.
Extensive experiments on FLORA-Bench show that GLOW consistently outperforms state-of-the-art baselines in both accuracy and ranking utility. 
Moreover, when integrated into the AFLOW framework, GLOW reduces the overall optimization time by 98.7\% with only a 0.031 average score decrease across three datasets, outperforming other performance prediction methods with substantially lower degradation.

In future work, we plan to extend our method so that it is not limited to outputting a binary success-based metric, which only measures whether an AW can produce the expected outcome. Specifically, we will incorporate additional dimensions such as execution cost (e.g., LLM invocation budget), runtime latency, and stability across stochastic executions, enabling a more practical and comprehensive multi-dimensional evaluation for AW.



\ifCLASSOPTIONcaptionsoff
\newpage
\fi



\normalem 
\bibliographystyle{IEEEtran}
\bibliography{mybibfile}
%



%

\vskip -0.5in
\begin{IEEEbiography}[{\includegraphics[width=1in,height=1.25in,clip,keepaspectratio]{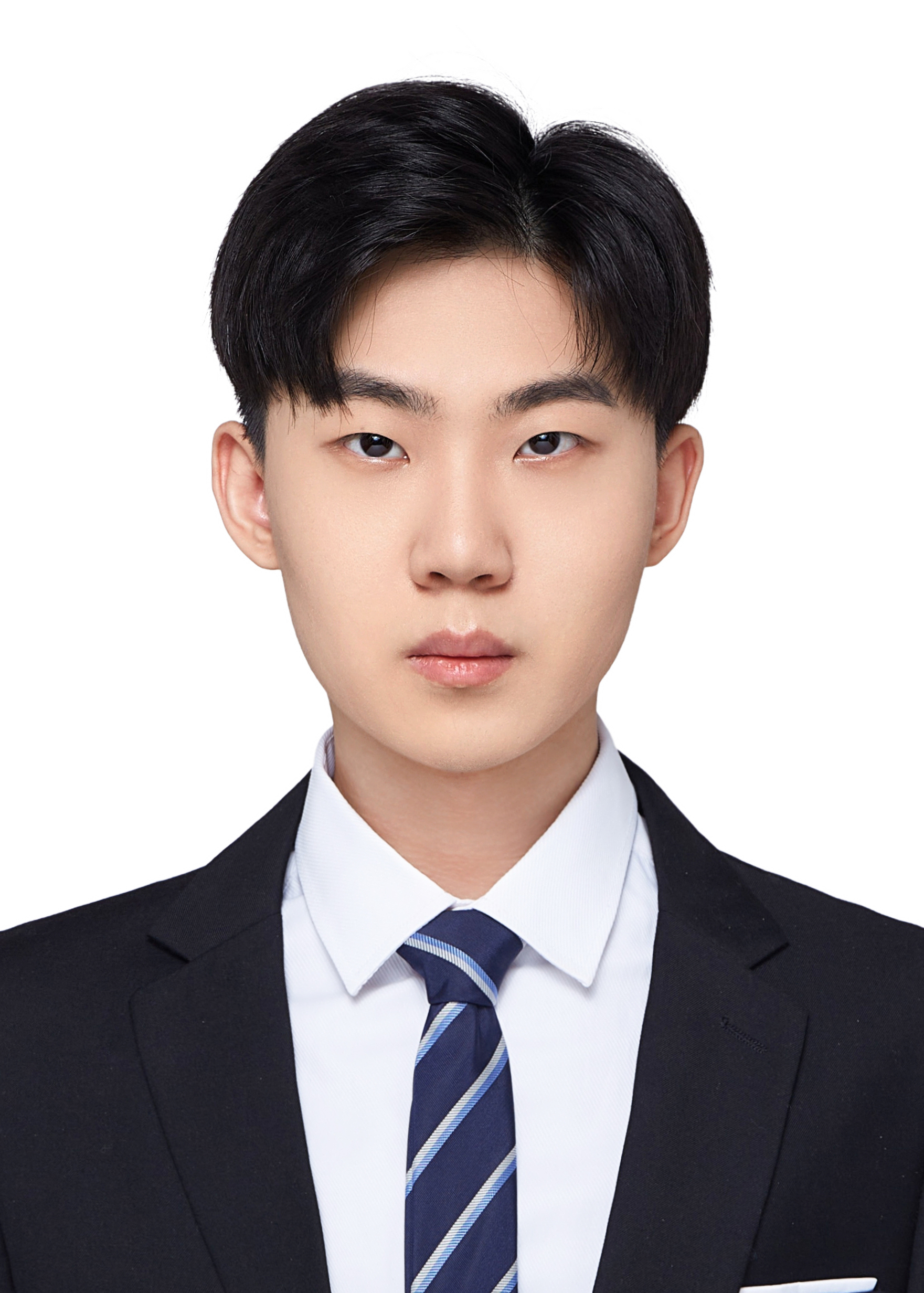}}]{Wei Guan}
    received the Ph.D. degree from Shanghai Jiao Tong University, China. He is currently an associate research fellow with the School of Computer Software, Tianjin University. His research interests include anomaly detection, large language models, and multi-agent systems.
\end{IEEEbiography}
	

\vskip -0.25in 
\begin{IEEEbiography}[{\includegraphics[width=1in,height=1.25in,clip,keepaspectratio]{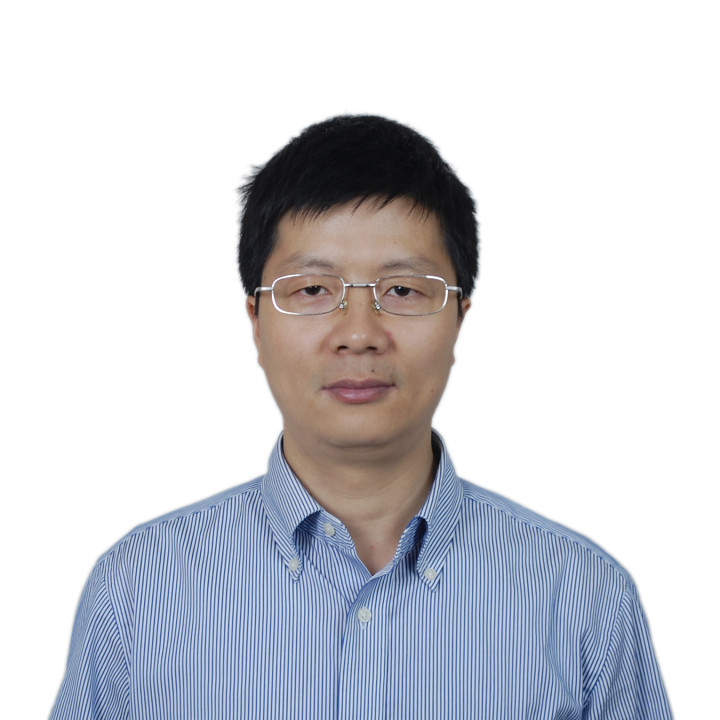}}]{Jian Cao}
    is currently a tenured professor with the School of Computer Science, Shanghai Jiao Tong University. He is also the director of research institute of network computing and service computing. Dr. Cao's research interests include intelligent data analytics, service computing.  He has published more than 300 research papers in prestigious journals and conferences. Currently, he is the distinguished member of CCF and the senior member of IEEE.
\end{IEEEbiography}

\vskip -0.25in
\begin{IEEEbiography}[{\includegraphics[width=1in,height=1.25in,clip,keepaspectratio]{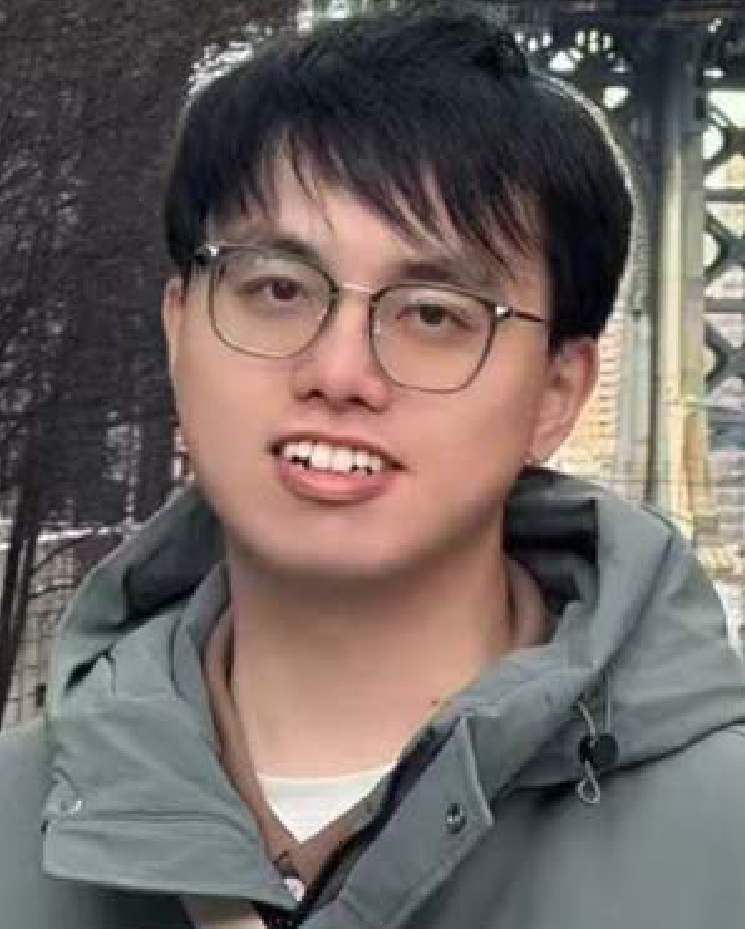}}]{Jinyu Cai} is currently a postdoctoral research fellow with the Institute of Data Science, National University of Singapore.   His research interests include anomaly/outlier detection, deep clustering, and graph machine learning.
\end{IEEEbiography}

\vskip -0.25in
\begin{IEEEbiography}[{\includegraphics[width=1in,height=1.25in,clip,keepaspectratio]{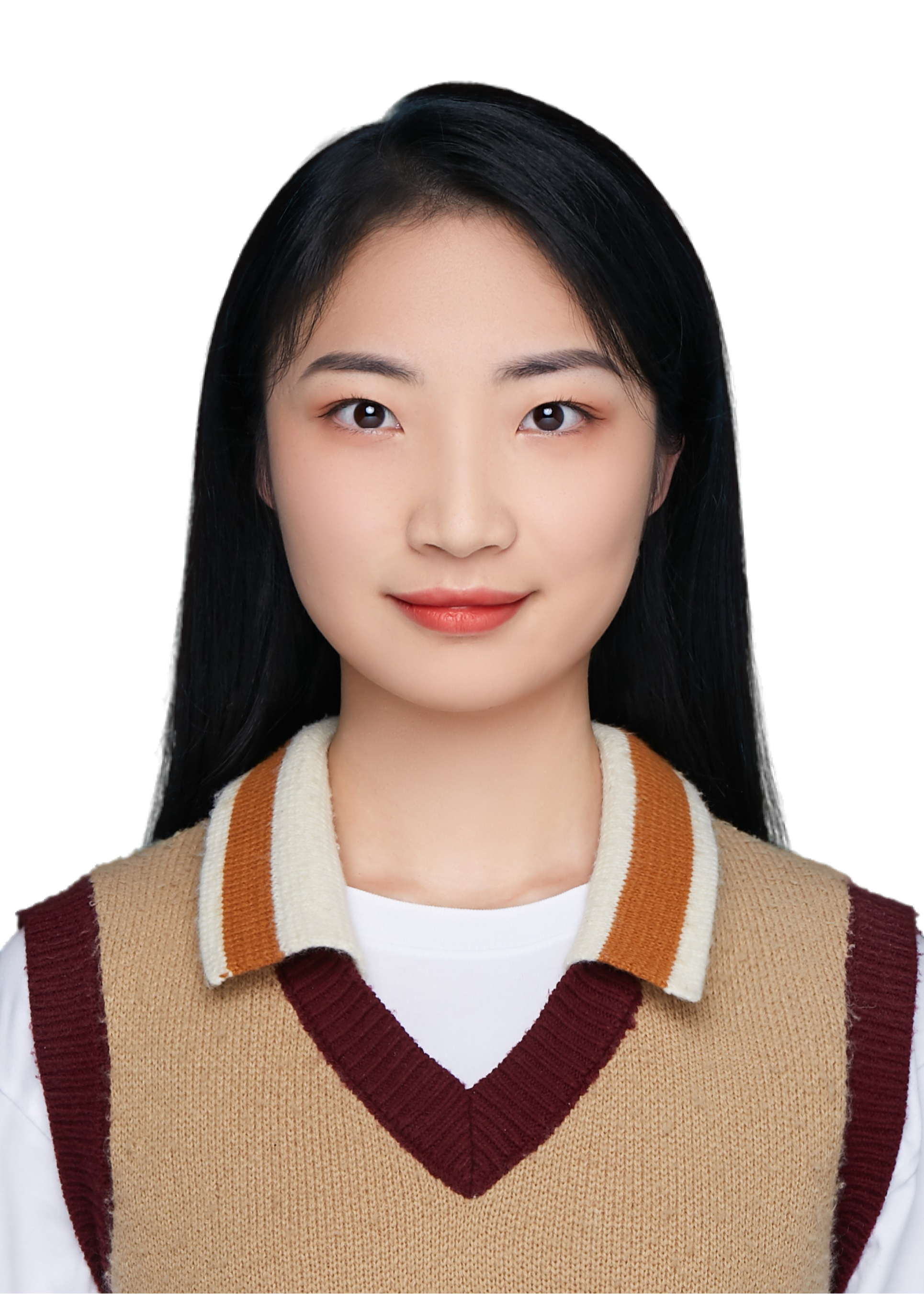}}]{Qiqi Cai}
    is currently a Ph.D student in the School of Computer Science, Shanghai Jiao Tong University. Her research interests include distributed AI, recommendation system and machine learning.
\end{IEEEbiography}

\vskip -0.25in
\begin{IEEEbiography}
[{\includegraphics[width=1in,height=1.25in,clip,keepaspectratio]{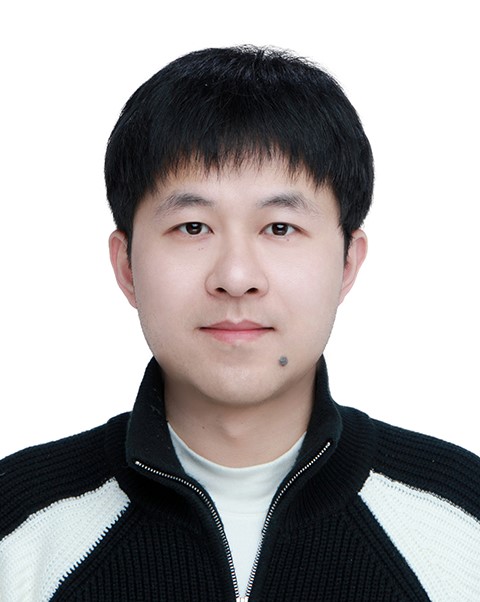}}]{Jianqi Gao}
is currently a lecturer with the School of Computer Engineering and Science, Shanghai University.  His research interests include event detection, adversarial learning, and knowledge engineering.
\end{IEEEbiography}

\vskip -0.25in
\begin{IEEEbiography}[{\includegraphics[width=1in,height=1.25in,clip,keepaspectratio]{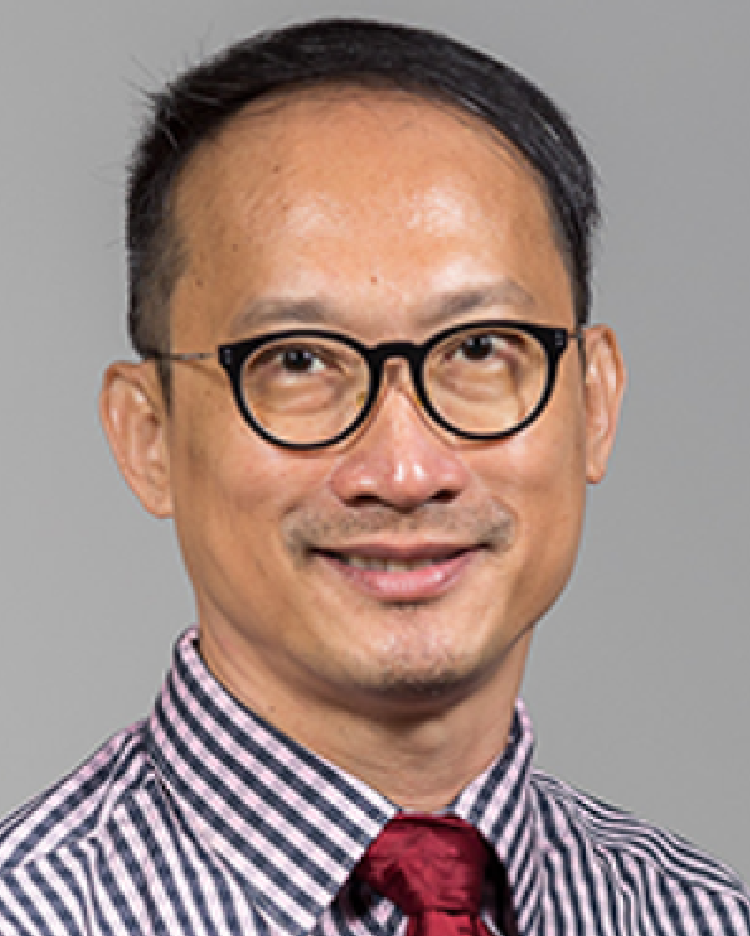}}]{See-Kiong Ng}
 is currently a professor of practice with the School of Computing, National University of Singapore, where he is also deputy director with the Institute of Data Science, as well as director of AI technology, AI Singapore. He has authored more than 130 papers on diverse and cross-disciplinary research topics, from bioinformatics to
smart cities, based on data science, and AI.
\end{IEEEbiography}




\clearpage

\end{document}